\pdfoutput=1
\documentclass[twoside]{article}
\usepackage{aistats2018}
\usepackage{amssymb}
 
\usepackage[draft]{hyperref}
\usepackage{pbox}
\usepackage{url}
\usepackage{algorithm}
\usepackage{algorithmic}
\usepackage{comment}
\usepackage{booktabs}       
\usepackage[tableposition=top]{caption}
\usepackage[titletoc,toc,title]{appendix}
\usepackage{subcaption}
\usepackage{amsfonts}       
\usepackage{nicefrac}       
\usepackage{microtype}      
\usepackage{graphicx}
\usepackage[raggedright]{sidecap}
\usepackage{natbib}
\usepackage{xspace}
\usepackage{todonotes}
\usepackage{amsmath}
\usepackage{amsopn}
\usepackage{multirow}

\usepackage{tikz}
\usetikzlibrary{bayesnet}
\newcommand{\pxz}{p_{\theta}(x|z)}
\newcommand{\qzx}{q_{\phi}(z|x)}
\DeclareMathOperator{\KL}{KL}
\newcommand{\E}{\gamma(z)}
\newcommand{\Exp}[2]{\mathop{\mathbb{E}}_{#2}\left[#1\right]}

\newcommand{\local}{\psi(x)}
\newcommand{\localopt}{\psi^*}
\newcommand{\MLP}{\text{MLP}}
\newcommand{\elbo}{\mathcal{L}(x;\theta,\psi(x))}
\newcommand{\optelbo}{\mathcal{L}(x;\theta,\psi^*)}
\newcommand{\Model}[2]{#1-#2}

\newcommand{\jacob}{\mathcal{J}(z)}

\begin{document}

\twocolumn[

\aistatstitle{On the challenges of learning with inference networks on sparse, high-dimensional data}
\aistatsauthor{ Rahul G. Krishnan \And Dawen Liang \And Matthew D. Hoffman }
\aistatsaddress{ MIT \And  Netflix \And Google} ]

\begin{abstract}
We study parameter estimation in Nonlinear Factor Analysis (NFA) where 
the generative model is parameterized by a deep neural network.
Recent work has focused on learning such models using inference (or recognition) networks;
we identify a crucial problem when modeling large, sparse, high-dimensional datasets -- underfitting.
We study the extent of underfitting, highlighting that its severity increases
with the sparsity of the data. We propose methods to tackle it via 
iterative optimization inspired by stochastic variational inference
\citep{hoffman2013stochastic} and improvements in the sparse data representation used for inference.
The proposed techniques drastically improve the ability of these powerful models to fit sparse data,
achieving state-of-the-art results on a benchmark text-count dataset
and excellent results on the task of top-N recommendation.
\end{abstract}

\section{Introduction}

Factor analysis \citep[FA,][]{spearman1904general} is a widely used latent variable model in the applied
sciences. The generative model assumes data, 
$x$, is a linear function of independent Normally distributed 
latent variables, $z$. The assumption of linearity has been relaxed 
in nonlinear factor analysis (NFA) \citep{gibson1960nonlinear} 
and extended across a variety of domains such as economics \citep{jones2006nonlinear}, 
signal processing \citep{jutten2003advances}, and machine learning \citep{valpola2002unsupervised, gpvlm}. 
NFA assumes the joint distribution factorizes as $p(x,z;\theta) = p(z) p(x|z; \theta)$. 
When the nonlinear conditional distribution is parameterized by deep neural networks
\citep[e.g.,][]{valpola2002unsupervised}, the graphical model
is referred to as a deep generative model. 
When paired with an inference or recognition network \citep{hinton1995wake}, 
a parametric function that approximates posterior distribution $p(z | x)$ (local variational parameters) from data,
such models go by the name of variational autoencoders \citep[VAEs,][]{kingma2013auto,rezende2014stochastic}.
We study NFA for the density estimation and analysis of \emph{sparse}, high-dimensional categorical data. 

Sparse, high-dimensional data is ubiquitous; it arises naturally in survey and demographic data,
bag-of-words representations of text, mobile app usage logs, recommender
systems, genomics, and electronic health records. 
Such data are characterized by
few frequently occurring features and a long-tail of rare features.
When directly learning VAEs on sparse data, a problem we run into
is that the standard learning algorithm results in \emph{underfitting} and fails to utilize the model's full capacity.
The inability to fully utilize the capacity of complex models in the large data regime is cause for
concern since it limits the applicability of this class of models to problems 
in the aforementioned domains. 

The contributions of this work are as follows. We identify a problem with standard VAE training when applied to sparse, high-dimensional data---underfitting.
We investigate the underlying causes of this phenomenon,
and propose modifications to the learning algorithm to address these causes. We combine inference networks with an 
iterative optimization scheme inspired by Stochastic Variational Inference (SVI) 
\citep{hoffman2013stochastic}. The proposed learning algorithm dramatically improves the quality of the estimated parameters.
We empirically study various factors that govern the severity of underfitting and how the techniques
we propose mitigate it. A practical ramification of our work is that improvements in learning NFA on recommender system data translate
to more accurate predictions and better recommendations. In contrast, standard VAE training fails to 
outperform the simple shallow linear models that still largely dominate the collaborative filtering domain \citep{effectiveLinear}.

\begin{SCfigure}[30][h]
	\begin{tikzpicture}[scale=0.75, transform shape,blackdot/.style={thin, draw=black, align=center, scale = 0.3,fill=black}]
	\node [latent] (z) {$z$};
	\node [obs,below= of z,yshift=-0.3cm] (x) {$x$};
	\draw [->] (z) to node[pos=0.45,fill=white] {$\theta$} (x);
	\node [latent, right=of z] (zq) {$z$};
	\node [obs,below= of z,right=of x] (xq) {$x$};
	\draw [->] (xq) to node[pos=0.45,fill=white] {$\phi$} (zq);
	\end{tikzpicture}
	\caption{\small \textbf{Nonlinear Factor Analysis: } {\bf [Left]} The generative model contains a single latent 
	variable $z$. 
	The conditional probability $p(x|z;\theta)$ parameterized by a deep neural network. 
	{\bf[Right]} The inference network $\qzx$ is used for inference at train and test time.}
	\label{fig:bn}
\end{SCfigure}

\section{Background\label{sec:background}}
\textbf{Generative Model: }
We consider learning in generative models of the form shown in Figure \ref{fig:bn}. We introduce the model
in the context of performing maximum likelihood estimation over a corpus of documents.
We observe a set of $D$ word-count\footnote{We use word-count in document for the sake of concreteness. 
Our methodology is generally applicable to other types of discrete high-dimensional data.} vectors $x_{1:D}$, where $x_{dv}$
denotes the number of times that word index $v\in\{1,\ldots,V\}$
appears in document $d$.  Given the total number of
words per document $N_d\equiv\sum_v x_{dv}$, $x_d$ is
generated via the following generative process:
\begin{align}
    \label{eqn:gen_model}
    &z_d \sim \mathcal{N}(0, I);\;
    \gamma(z_d) \equiv \MLP(z_d; \theta);\;\\
    &\mu(z_d) \equiv \frac{\exp\{\gamma(z_d)\}}{\sum_v \exp\{\gamma(z_d)_{v}\}};\;
    x_d \sim \mathrm{Mult.}(\mu(z_d), N_d).\nonumber
\end{align}
That is, we draw a Gaussian random vector, pass it through a
multilayer perceptron (MLP) parameterized by $\theta$, pass the
resulting vector through the softmax (a.k.a. multinomial logistic)
function, and sample $N_d$ times from the resulting distribution over
$V$.\footnote{In keeping with common practice, we neglect
the multinomial base measure term $\frac{N!}{x_1!\cdots x_V!}$, which
amounts to assuming that the words are observed in a particular
order.}

\textbf{Variational Learning: }
For ease of exposition we drop the subscript on $x_d$ when referring to a single data point.
Jensen's inequality yields the following lower bound on the log marginal likelihood of the data: 

\begin{equation}
\begin{split}
\label{eqn:varlowbnd}
\log p_\theta(x)\ge
\underbrace{\mathbb{E}_{q(z; \psi)}[\log p_\theta(x\mid z)] - \KL(\, q(z; \psi)\mid\mid p(z)\, ).}_{
\mathcal{L}(x; \theta, \psi)}
\end{split}
\end{equation}
$q(z;\psi)$ is a tractable ``variational'' distribution meant to
approximate the intractable posterior distribution $p(z\mid x)$; it is
controlled by some parameters $\psi$.  For example, if $q$ is
Gaussian, then we might have $\psi = \{\mu, \Sigma\}$, $q(z;\psi) =
\mathcal{N}(z;\mu, \Sigma)$.
We are free to choose $\psi$ however we want, but ideally we would choose
the $\psi$ that makes the bound in equation \ref{eqn:varlowbnd} as tight
as possible, $\psi^* \triangleq \arg \max_\psi \mathcal{L}(x; \theta, \psi)$.

\citet{hoffman2013stochastic} proposed finding $\psi^*$ using
iterative optimization, starting from a random initialization. This is
effective, but can be costly. More recently, \citet{kingma2013auto}
and \citet{rezende2014stochastic} proposed training a feedforward
\emph{inference network} \citep{hinton1995wake} to find good
variational parameters $\psi(x)$ for a given $x$, where $\psi(x)$ is
the output of a neural network with parameters $\phi$ that are trained
to maximize $\mathcal{L}(x; \theta, \psi(x))$. Often it is much
cheaper to compute $\psi(x)$ than to obtain an optimal $\psi^*$ using
iterative optimization. But there is no guarantee that $\psi(x)$
produces optimal variational parameters---it may yield a much looser
lower bound than $\psi^*$ if the inference network is either not
sufficiently powerful or its parameters $\phi$ are not well tuned.

Throughout the rest of this paper, we will use $\psi(x)$ to denote an
inference network that implicitly depends on some parameters $\phi$,
and $\psi^*$ to denote a set of variational parameters obtained by
applying an iterative optimization algorithm to equation
\ref{eqn:varlowbnd}. Following common convention, we will sometimes
use $q_\phi(z\mid x)$ as shorthand for $q(z; \psi(x))$.

\section{Mitigating Underfitting\label{sec:underfitting}}
We elucidate our hypothesis on why the learning algorithm for VAEs is susceptible to underfitting.

\subsection{Tight Lower Bounds on $\log p(x)$} 

There are two sources of error in variational parameter estimation with inference networks:

The first is the distributional error accrued due to learning with a tractable-but-approximate family of distributions $\qzx$
instead of the true posterior distribution $p(z|x)$. Although difficult to compute in practice, 
it is easy to show that this error is exactly $\KL(\qzx \| p(z|x))$. We restrict ourselves
to working with normally distributed variational approximations and do not aim to overcome this source of error. 

\begin{figure}[h]
	\centering
	\includegraphics[width=0.5\linewidth]{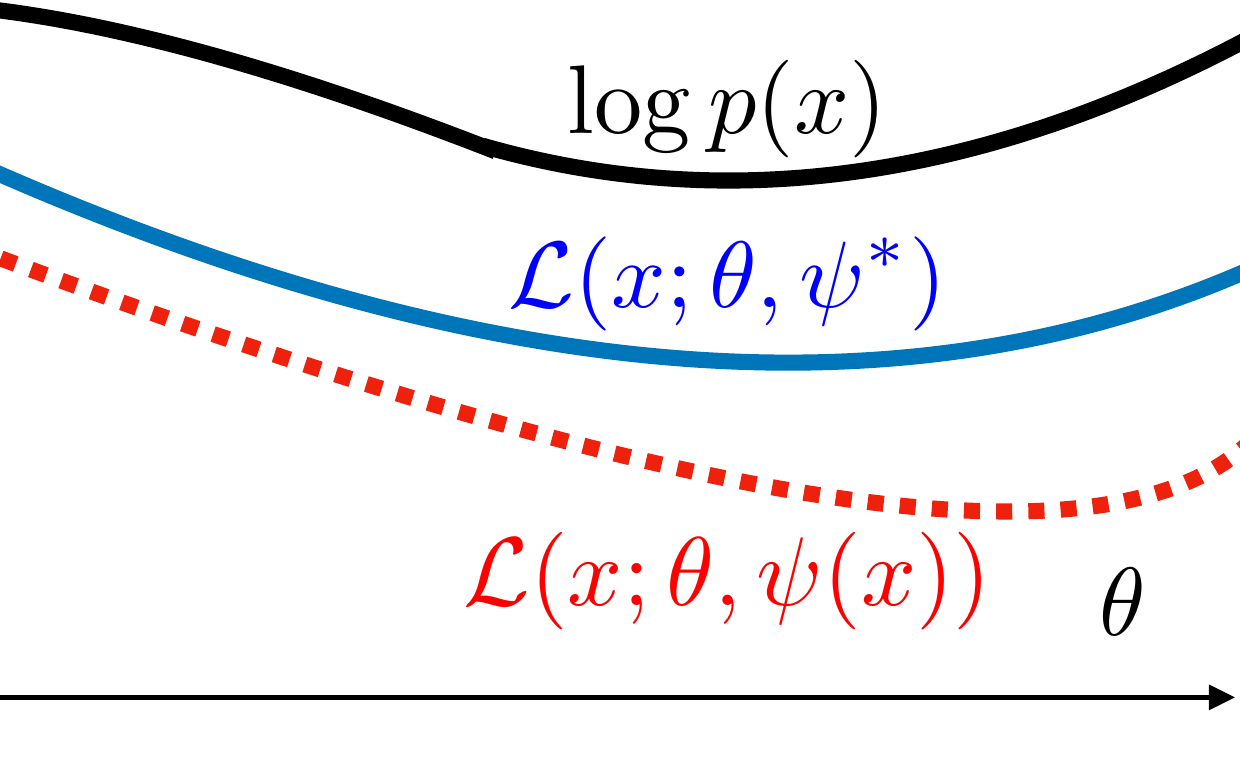}
\caption{\small
\textbf{Lower Bounds in Variational Learning: } 
To estimate $\theta$, we maximize a lower bound on $\log p(x;\theta)$. 
$\elbo$ denotes the standard training
objective used by VAEs. The tightness of this bound (relative
to $\optelbo$ depends on the inference network. The x-axis is $\theta$. 
}
\label{fig:lower_bounds}
\end{figure}

The second source of error comes from the sub-optimality of the
variational parameters $\psi$ used in Eq. \ref{eqn:varlowbnd}. We are guaranteed 
that $\elbo$ is a valid lower bound on $\log p(x)$ for \emph{any} output of $\qzx$ but within the same
family of variational distributions, there exists an optimal choice of variational parameters $\psi^*=\{\mu^*,\Sigma^*\}$
realizing the tightest variational bound for a data point $x$.  
\begin{align}
	\label{eqn:opt_local}
	&\log p(x) \geq \\
	&\underbrace{\mathbb{E}_{\mathcal{N}(\mu^*;\Sigma^*)}[\log p(x|z;\theta)]-\KL(\mathcal{N}(\mu^*,\Sigma^*) \| p(z))}_{\optelbo}\nonumber\\
	&\geq \elbo\nonumber
\end{align}
\begin{align}
	&\text{Where we define: } \psi^* := \{\mu^*,\Sigma^*\}\nonumber\\
	&= \arg\max_{\mu,\Sigma} \mathbb{E}_{\mathcal{N}(\mu,\Sigma)}[\log p(x|z;\theta)] - \KL(\mathcal{N}(\mu,\Sigma) \| p(z)).\nonumber
\end{align}
The cartoon in Fig. \ref{fig:lower_bounds} illustrates this double bound.

Standard learning algorithms for VAEs update $\theta,\phi$ jointly based on $\elbo$ (see Alg. \ref{alg:vae_update} for pseudocode).
That is, they directly use $\local$ (as output by $\qzx$) to estimate Equation \ref{eqn:varlowbnd}.

In contrast, older stochastic variational inference methods
\citep{hoffman2013stochastic} update $\theta$ based on gradients of $\mathcal{L}(x;\theta,\psi^*)$
by updating randomly initialized variational parameters for each example. $\localopt$ is obtained by maximizing $\mathcal{L}(x;\theta,\psi)$ with respect to $\psi$.
This maximization is performed by $M$ gradient ascent steps yielding $\psi_M\approx\psi^*$.
(see Alg. \ref{alg:svi_update}). 

\begin{algorithm}[h]
\begin{algorithmic}
\STATE \textbf{Inputs}: $\mathcal{D}:=\left[x_1,\ldots,x_D\right],$\\
Model: $\qzx$, $\pxz$, $p(z)$;
\FOR{k = 1\ldots K}
\STATE Sample: $x\sim\mathcal{D}$, $\local = \qzx$, update $\theta,\phi$:\\
\qquad $\theta^{k+1}\leftarrow \theta^k + \eta_{\theta}\nabla_{\theta^k}\mathcal{L}(x;\theta^k,\psi(x))$\\
\qquad $\phi^{k+1}\leftarrow\phi^k+\eta_{\phi}\nabla_{\phi^k}\mathcal{L}(x;\theta^{k},\psi(x))$
\ENDFOR
\end{algorithmic}
\caption{\small \label{alg:vae_update} \textbf{Learning with Inference Networks} \citep{kingma2014semi}
}
\end{algorithm}

\begin{algorithm}[h]
\begin{algorithmic}
\STATE \textbf{Inputs}: $\mathcal{D}:=\left[x_1,\ldots,x_D\right],$\\
Model: $\pxz$, $p(z)$;
\FOR{k = 1\ldots K}
\STATE {\bf 1.} Sample: $x\sim\mathcal{D}$ and initialize: $\psi_0 = \mu_0,\Sigma_0$
\STATE {\bf 2.} Approx. $\psi_M\approx\psi^*= \arg\max_{\psi} \mathcal{L}(x;\theta;\psi)$:\\
\qquad For $m=0,\ldots,M-1$:\\
	\qquad\qquad $\psi_{m+1} = \psi_{m}+\eta_{\psi}\frac{\partial \mathcal{L}(x;\theta^k,\psi_m)}{\partial \psi_{m}}$
\STATE {\bf 3.} Update $\theta$:
$\theta^{k+1}\leftarrow \theta^k + \eta_{\theta}\nabla_{\theta^k}\mathcal{L}(x;\theta^k,\psi_{M})$
\ENDFOR
\end{algorithmic}
\caption{\small \label{alg:svi_update} \textbf{Learning with Stochastic Variational Inference}: 
$M$: number of gradient updates to $\psi$. 
}
\end{algorithm}

\subsection{Limitations of Joint Parameter Updates}
Alg. \eqref{alg:vae_update} updates $\theta, \phi$ jointly.
During training, the inference network learns to approximate the posterior, 
and the generative model improves
itself using local variational parameters $\local$ output by $\qzx$.

If the variational parameters $\psi(x)$ output by 
the inference network are close to the optimal variational 
parameters $\psi^*$ (Eq. \ref{eqn:opt_local}), then the updates
for $\theta$ are based on a relatively tight lower bound on $\log p(x)$.
But in practice $\psi(x)$ may not be a good approximation to $\psi^*$.

Both the inference network and generative model are initialized randomly. 
At the start of learning, $\local$ is the output of a randomly initialized neural network,
and will therefore be a poor approximation to the optimal parameters $\psi^*$.
So the gradients used to update $\theta$ will be based on a very loose lower bound on $\log p(x)$.
These gradients may push the generative model towards a poor local minimum---
previous work has argued that deep neural networks (which form the conditional probability distributions $\pxz$) 
are often sensitive to initialization
\citep{glorot2010understanding, larochelle2009exploring}.
Even later in learning, $\psi(x)$ may yield suboptimal gradients for $\theta$ if the inference
network is not powerful enough to find optimal variational parameters for all data points.

Learning in the original SVI scheme does not suffer from this problem,
since the variational parameters are optimized within the inner loop of 
learning before updating to $\theta$ (i.e. in Alg. \eqref{alg:svi_update}; $\partial\theta$ is effectively derived using $\optelbo$).
However, this method requires potentially an expensive iterative optimization.

This motivates blending the two methodologies for parameter estimation.
Rather than rely entirely on the inference network, we use its output
to ``warm-start'' an SVI-style optimization that yields higher-quality
estimates of $\psi^*$, which in turn should yield more meaningful gradients
for $\theta$.

\subsection{Optimizing Local Variational Parameters \label{sec:method}}
We use the local variational parameters $\psi=\local$ predicted by the inference network to initialize an
iterative optimizer. 
As in Alg. \ref{alg:svi_update}, we perform gradient ascent 
to maximize $\mathcal{L}(x;\theta,\psi)$ with respect to $\psi$.
The resulting $\psi_M$ approximates the optimal variational parameters:  $\psi_M\approx\psi^*$.
Since NFA is a continuous latent variable model, these updates can be achieved via 
the re-parameterization gradient \citep{kingma2013auto}.
We use $\psi^*$ to derive gradients for $\theta$ under $\optelbo$.
Finally, the parameters of the inference network ($\phi$) are updated using stochastic backpropagation
and gradient descent, holding fixed the parameters of the generative model ($\theta$). 
Our procedure is detailed in Alg. \ref{alg:var_opt}.

\begin{algorithm}[h]
\begin{algorithmic}
\STATE \textbf{Inputs}: $\mathcal{D}:=\left[x_1,\ldots,x_D\right],$\\
Inference Model: $\qzx$, \\
Generative Model: $\pxz,p(z)$,
\FOR{k = 1\ldots K}
\STATE {\bf 1.} Sample: $x\sim\mathcal{D}$ and set $\psi_0=\psi(x)$
\STATE {\bf 2.} Approx. $\psi_M\approx\psi^*= \arg\max_{\psi} \mathcal{L}(x;\theta^k;\psi)$,\\
\qquad For $m=0,\ldots,M-1$:\\
	\qquad\qquad $\psi_{m+1} = \psi_{m}+\eta_{\psi}\frac{\partial \mathcal{L}(x;\theta^k,\psi_m)}{\partial \psi_{m}}$
\STATE {\bf 3.} Update $\theta$,\\
\qquad $\theta^{k+1}\leftarrow \theta^k + \eta_{\theta}\nabla_{\theta^k}\mathcal{L}(x;\theta^k,\psi_{M})$
\STATE {\bf 4.} Update $\phi$,\\
\qquad $\phi^{k+1}\leftarrow\phi^k+\eta_{\phi}\nabla_{\phi^k}\mathcal{L}(x;\theta^{k+1},\psi(x))$
\ENDFOR
\end{algorithmic}
\caption{\small \label{alg:var_opt} \textbf{Maximum Likelihood Estimation of $\theta$ with Optimized Local Variational Parameters:} 
Expectations in $\mathcal{L}(x,\theta,\psi^*)$ (see Eq. \ref{eqn:opt_local}) are evaluated with a single
sample from the optimized variational distribution. 
$M$ is the number of updates to the variational parameters ($M=0$ implies no additional optimization).
$\theta,\local,\phi$ are updated using stochastic gradient descent with
learning rates $\eta_{\theta},\eta_{\psi},\eta_{\phi}$ obtained via ADAM \citep{kingma2014adam}.
In step $4$, we update $\phi$ separately from $\theta$. One could 
alternatively, update $\phi$ using $\KL(\local_M \| \qzx)$ as in \citet{salakhutdinov2010efficient}. 
}
\end{algorithm}

\subsection{Representations for Inference Networks}
The inference network must learn to regress to the 
optimal variational parameters for any combination of
features, but in sparse datasets, many words appear only rarely.
To provide more global context about rare words, we provide to the inference network (but not
the generative network) TF-IDF \citep{baeza1999modern} features
instead of raw counts. These give the inference network a hint
that rare words are likely to be highly informative.
TF-IDF is a popular technique in information retrieval that re-weights features to increase the influence of 
rarer features while decreasing the influence of common features. 
The transformed feature-count vector is
  $\tilde x_{dv} \equiv x_{dv}\log\frac{D}{\sum_{d'} \min\{x_{d'v}, 1\}}$.
The resulting vector $\tilde x$ is then normalized by its L2 norm.

\section{Related Work}
\citeauthor{salakhutdinov2010efficient} optimize local mean-field parameters from an inference network in the context
of learning deep Boltzmann machines. 
\citeauthor{salimans2015markov} explore warm starting MCMC with the output of an inference network. 
\citeauthor{hjelm2015iterative} explore a similar idea as ours to derive an importance-sampling-based 
bound for learning deep generative models with discrete latent variables. 
They find that learning with $\psi^*$ does not improve results on binarized MNIST.
This is consistent with our experience---we find that our secondary
optimization procedure helped more when learning models of sparse data.
\citeauthor{miao2015neural} learn log-linear models \citep[multinomial-logistic PCA,][]{collins2001generalization} 
of documents using inference networks. We show that mitigating underfitting in deeper models
yields better results on the benchmark RCV1 data.
Our use of the spectra of the Jacobian matrix of $\log p(x|z)$ to inspect learned models 
is inspired by \citet{wang2016analysis}. 

Previous work has studied the failure modes of learning VAEs. They can be broadly categorized into two classes.
The first aims to improves the utilization of latent variables using a richer posterior distribution \citep{burda2015importance}.
However, for sparse data, the limits of learning with a Normally distributed $\qzx$ have barely been pushed -- our goal is to do so in this work.
Further gains may indeed be obtained with a richer posterior distribution but the techniques
herein can inform work along this vein. 
The second class of methods studies ways to alleviate the underutilization of latent dimensions due to an overtly expressive choice of 
models for $p(x|z;\theta)$ such as a Recurrent Neural Network \citep{bowman2015generating, chen2016variational}.
This too, is not the scenario we are in; underfitting of VAEs on sparse data occurs even when $p(x|z;\theta)$ is an MLP.

Our study here exposes a third failure mode; one in which learning is challenging not just because of the objective used in learning
but also because of the \emph{characteristics} of the data.

\section{Evaluation\label{sec:evaluation}}

We first confirm our hypothesis empirically that underfitting is an issue when learning VAEs on high dimensional sparse datasets. 
We quantify the gains (at training and test time) obtained by the use of TF-IDF features and the continued optimization of 
$\local$ on two different types of high-dimensional sparse data---text and movie ratings.
In Section \ref{sec:bow}, we learn VAEs on two large scale bag-of-words datasets. 
We study (1) \emph{where} the proposed methods might have the most impact and (2) present evidence for \emph{why} the learning 
algorithm (Alg. \ref{alg:var_opt}) works.
In Section \ref{sec:recsys}, we show that improved inference is crucial
to building deep generative models that can tackle problems in top-$N$ recommender systems.
We conclude with a discussion. 

\subsection{Setup} 

\textbf{Notation: } 
In all experiments, 
$\local$ denotes learning with Alg. \ref{alg:vae_update} and 
$\localopt$ denotes the results of learning with Alg. \ref{alg:var_opt}.
$M=100$ (number of updates to the local variational parameters) on the bag-of-words text data
and $M=50$ on the recommender systems task. 
$M$ was chosen based on the number of steps it takes for $\mathcal{L}(x;\theta,\psi_m)$ (Step $2$ in Alg. \ref{alg:var_opt}) to converge
on training data.
\Model{3}{$\psi^*$}-norm denotes a model where the MLP parameterizing $\E$ has three layers: two hidden layers and one output layer,
$\psi^*$ is used to derive an update of $\theta$ and normalized count features are conditioned on by the inference network.
In all tables, we display evaluation metrics obtained under 
both $\local$ (the output of the inference network) and $\psi^*$ (the optimized variational parameters). 
In figures, we always display metrics obtained under $\psi^*$ (even if the model was trained 
with $\local$) since $\optelbo$ always forms a tighter bound to $\log p(x)$. 
If left unspecified TF-IDF features are used as input to the inference network. 

\textbf{Training and Evaluation: } 
We update $\theta$ using learning rates given by ADAM \citep{kingma2014adam} (using a batch size of $500$),
The inference network's intermediate hidden layer $h(x)=\text{MLP}(x;\phi_0)$ (we use a two-layer MLP in the inference network for all experiments)
are used to parameterize the mean and diagonal log-variance as:  
$\mu(x) = W_{\mu}h(x),\log\Sigma(x)=W_{\log\Sigma}h(x)$ where $\phi = \{W_{\mu},W_{\log\Sigma},\phi_0\}$. 
Code is available at \url{github.com/rahulk90/vae_sparse}.

\subsection{Bag-of-words text data\label{sec:bow}}

\textbf{Datasets and Metrics: }  
We study two large text datasets. 
(1) RCV1 \citep{lewis2004rcv1} dataset (train/valid/test: 789,414/5,000/10,000, $V$: 10,000).
We follow the preprocessing procedure in \citet{miao2015neural}. 
(2) The Wikipedia corpus used in \citet{HuangEtAl2012} (train/test: 1,104,937/100,000 and $V$:20,000).
We set all words to lowercase, ignore numbers and restrict the dataset to the top $20,000$ frequently occurring words. 
We report an upper bound on perplexity \citep{mnih2014neural} given 
by $\exp(-\frac{1}{N}\sum_{i}\frac{1}{N_i}\log p(x_i))$ where $\log p(x_i)$ is replaced by Eq \ref{eqn:varlowbnd}.
To study the utilization of the latent dimension obtained by various training methods,  
we compute the Jacobian $\jacob$ matrix (as $\nabla_z \log p(x|z)$). 
The singular value spectrum of the Jacobian directly measures the \emph{utilization} of the latent dimensions in the model.
We provide more details on this in the supplementary material.

\begin{figure*}[ht!]
\centering
\begin{subfigure}{.3\linewidth}
  \centering
  \includegraphics[width=\linewidth]{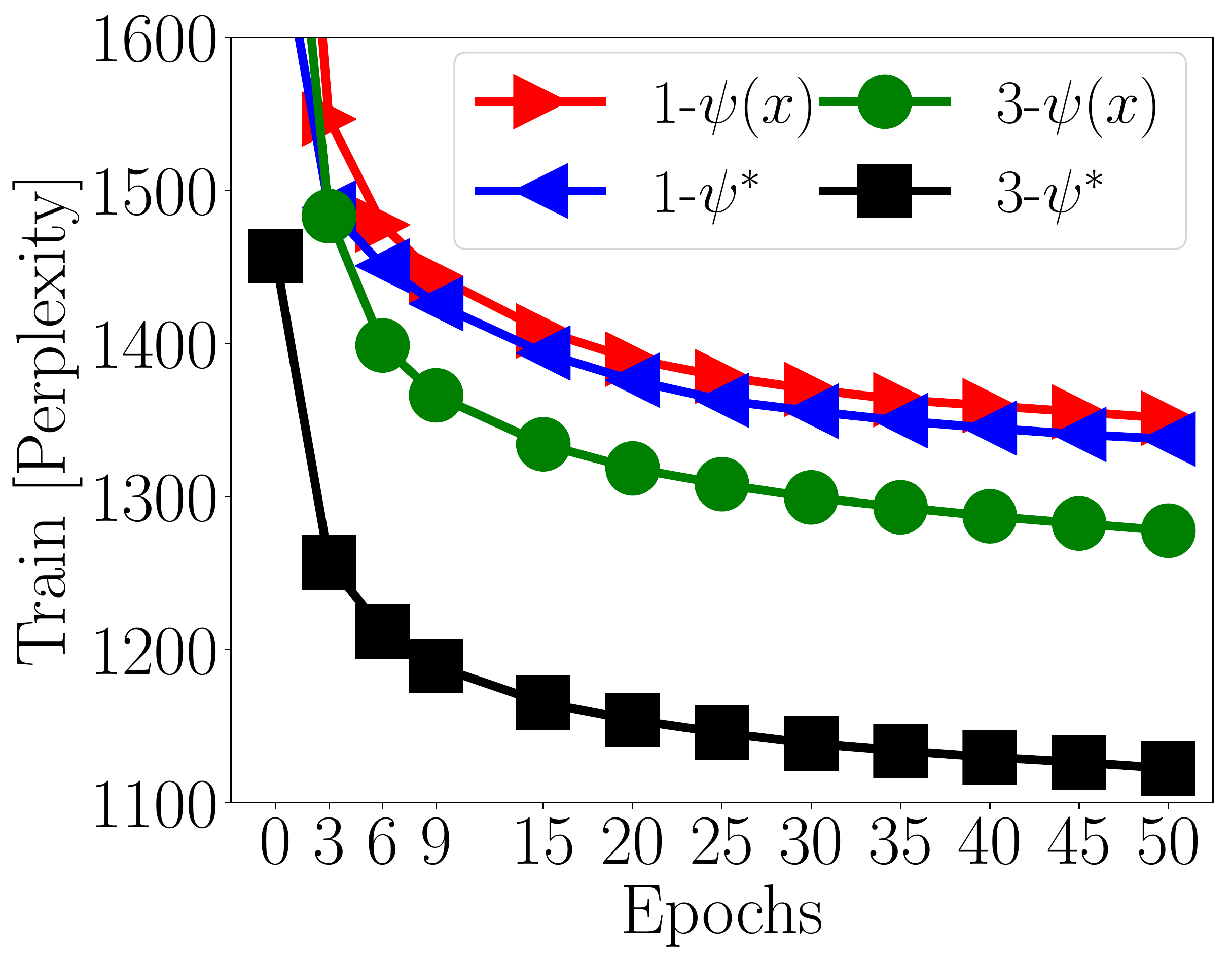}
  \caption{Wikipedia - Training}
  \label{fig:wiki_train_bounds}
\end{subfigure}\begin{subfigure}{.3\linewidth}
  \includegraphics[width=0.97\linewidth]{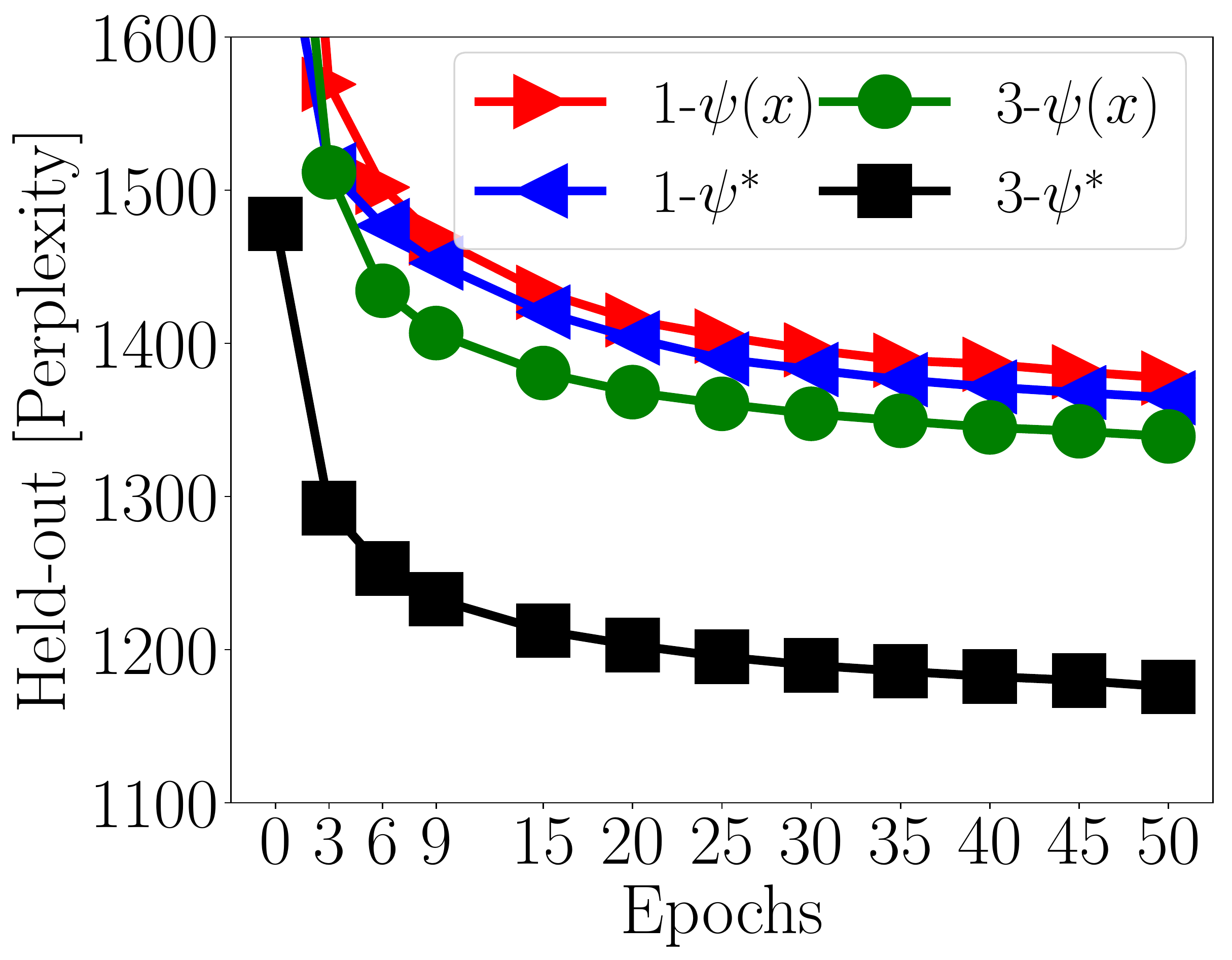}
\caption{Wikipedia - Evaluation}
\label{fig:wiki_eval_bounds}
\end{subfigure}\begin{subfigure}{.38\linewidth}
	\includegraphics[width=\linewidth]{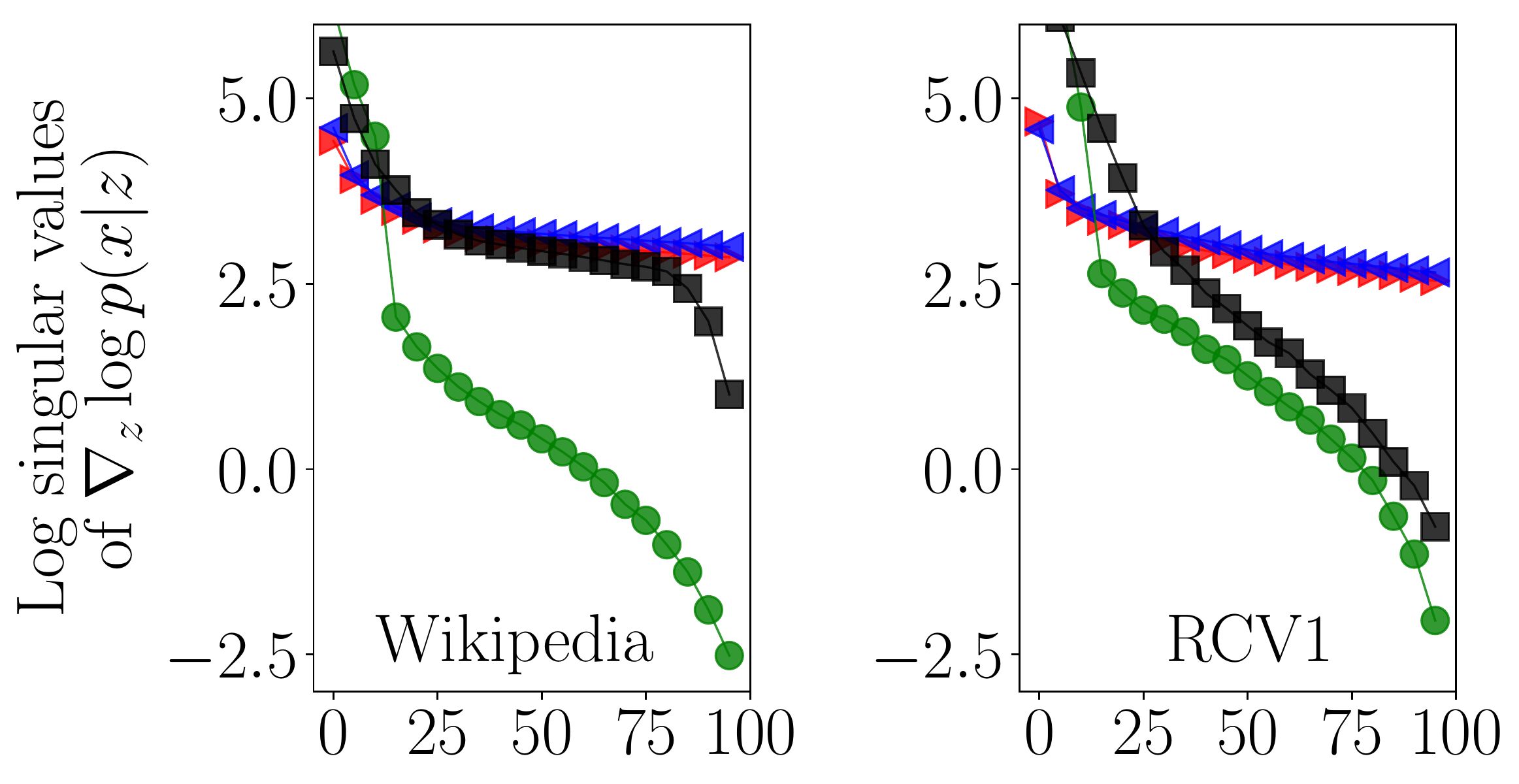}
\caption{Log-Singular Values}
\label{fig:wiki_logsingular}
\end{subfigure}
\caption{
\small
\textbf{Mechanics of Learning: }
Best viewed in color. 
\textbf{(Left and Middle)} For the Wikipedia dataset, we visualize 
upper bounds on training and held-out perplexity (\emph{evaluated} with $\psi^*$) 
viewed as a function of epochs. 
Items in the legend corresponds to choices of training method. 
\textbf{(Right)} Sorted log-singular values of $\nabla_z\log p(x|z)$ on Wikipedia (left)
on RCV1 (right) for different training methods. The x-axis is latent dimension.
The legend is identical to that in Fig. \ref{fig:wiki_train_bounds}.
}
\label{fig:bounds}
\end{figure*}

\begin{table}[b!]
\caption{\small 
\textbf{Test Perplexity on RCV1: } 
\textbf{Left: Baselines} 
Legend: LDA \citep{blei2003latent}, 
Replicated Softmax (RSM) \citep{hinton2009replicated},
Sigmoid Belief Networks (SBN) and 
Deep Autoregressive Networks (DARN) \citep{mnih2014neural}, 
Neural Variational Document Model (NVDM) \citep{miao2015neural}.
$K$ denotes the latent dimension in our notation.
\textbf{Right: } NFA on text data with $K=100$.  We vary the features presented
to the inference network $\qzx$ during learning between: 
normalized count vectors ($\frac{x}{\sum_{i=1}^{V} x_i}$, denoted ``norm'') 
and normalized TF-IDF }	\label{tab:perplexity}
	\resizebox{\linewidth}{!}{
		\centering
		\begin{tabular}{cc|c}
			\multicolumn{1}{c}{\bf Model}  &\multicolumn{1}{c}{\bf $K$} &\multicolumn{1}{c}{\bf RCV1}\\
			\hline 
			LDA &50 &1437\\
			LDA &200 &1142\\
			RSM &50 &988\\
			\hline
			SBN &50 &784\\
			fDARN &50 &724\\
			fDARN &200 &598\\
			\hline
			NVDM &50 &563\\
			NVDM &200 &550\\
		\end{tabular}
					\centering
		\begin{tabular}{c|cc}
                  {\bf NFA} 
      &  $\local$ & $\psi^*$\\
      \hline
      \Model{1}{$\local$}-norm  & 501 & 481\\
      \Model{1}{$\psi^*$}-norm   & 488 & 454\\
      \Model{3}{$\local$}-norm   & 396 & 355\\
      \Model{3}{$\psi^*$}-norm  & 378 & {\bf 331}\\
      \hline
      \Model{1}{$\local$}-tfidf   & 480 & 456\\
      \Model{1}{$\psi^*$}-tfidf & 482 & 454\\
      \Model{3}{$\local$}-tfidf & 384 & 344 \\
      \Model{3}{$\psi^*$}-tfidf & 376 & \textbf{331}\\
    \end{tabular}
    }
\end{table}

\textbf{Reducing Underfitting: }
Is underfitting a problem and does optimizing $\local$ with the use of TF-IDF features help? 
Table \ref{tab:perplexity} confirms both statements. 

We make the following observations:
(1) between ``norm'' and ``tfidf'' (comparing first four rows and second four rows), 
we find that the use of TF-IDF features almost always improves parameter estimation;
(2) optimizing $\local$ at test time (comparing column $\psi^*$ with $\local$) always yields a tighter bound on $\log p(x)$,
often by a wide margin. Even after extensive training the inference
network can fail to tightly approximate $\optelbo$, suggesting that there may be limitations to the power of generic amortized inference;
(3) optimizing $\local$ during training ameliorates under-fitting and yields significantly
better generative models on the RCV1 dataset. 
We find that the degree of underfitting and subsequently the improvements from training with $\psi^*$ 
are significantly more pronounced on the larger and sparser 
Wikipedia dataset (Fig. \ref{fig:wiki_train_bounds} and \ref{fig:wiki_eval_bounds}). 

\textbf{Effect of optimizing $\local$: }
How does learning with $\localopt$ affect the rate of convergence the learning algorithm?
We plot the upper bound on perplexity versus epochs
on the Wikipedia (Fig. \ref{fig:wiki_train_bounds}, \ref{fig:wiki_eval_bounds}) datasets. 
As in Table \ref{tab:perplexity}, the additional optimization does not appear to help much when the generative
model is linear. On the deeper three-layer model, 
learning with $\psi^*$ dramatically improves the model allowing it to fully utilize its potential for density estimation.
Models learned with $\psi^*$ quickly converge to a better local minimum
early on (as reflected in the perplexity evaluated on the training data and held-out data).
We experimented with continuing to train \Model{3}{$\local$} beyond $150$ epochs, where it reached a validation perplexity of 
approximately $1330$, worse than that obtained by \Model{3}{$\psi^*$} at epoch $10$ suggesting that longer training is unsufficient
to overcome local minima issues afflicting VAEs. 

\textbf{Overpruning of latent dimensions: }
One cause of underfitting is due to overpruning of the latent dimensions in the model.
If the variational distributions for a subset of the latent dimensions
of $z$ are set to the prior, this effectively reduces the model's capacity. 
If the KL-divergence in Eq. \ref{eqn:varlowbnd} encourages the approximate posterior to remain 
close to the prior early in training, \emph{and} if
the gradient signals from the likelihood term are weak or inconsistent, the KL 
may dominate and prune out latent dimensions before the model can use them.

In Fig. \ref{fig:wiki_logsingular}, we plot the log-spectrum of the Jacobian matrices for different training methods and models.
For the deeper models, optimizing $\local$ is crucial to utilizing its 
capacity, particularly on the sparser Wikipedia data. Without it, only about ten latent
dimensions are used, and the model severely underfits the data. 
Optimizing $\local$ iteratively likely limits overpruning since the variational parameters ($\psi^*$) don't solely focus
on minimizing the KL-divergence but also on maximizing the likelihood of the data (the first term 
in Eq. \ref{eqn:varlowbnd}). 
\begin{figure}[h]
	\centering
  \includegraphics[width=0.8\linewidth]{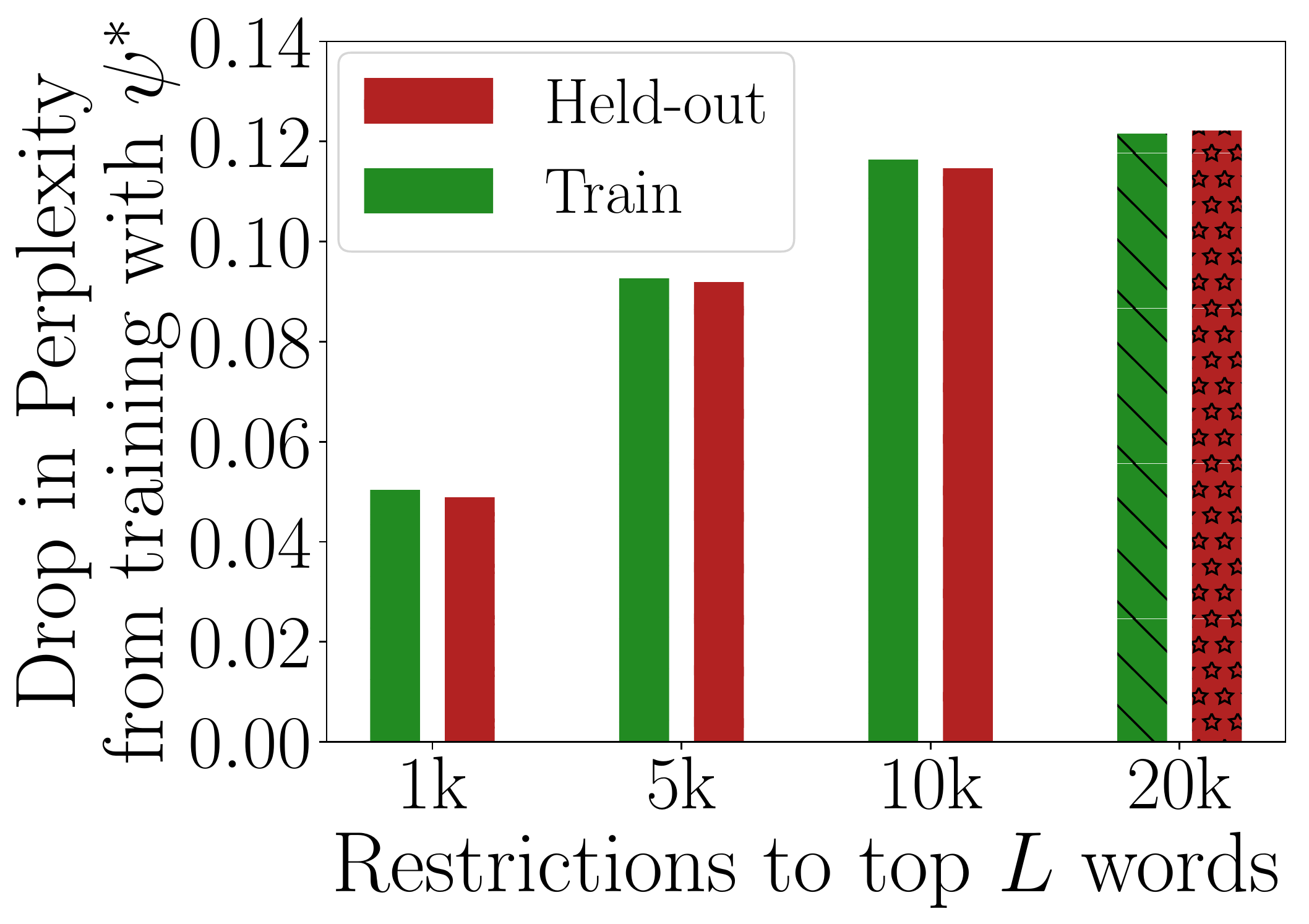}
\caption{
\small
\textbf{Decrease in Perplexity versus Sparsity: }
We plot the relative drop in perplexity obtained by training with $\psi^*$
instead of $\local$ against varying levels of sparsity in the Wikipedia data.
On the y-axis, we plot $\frac{P_{[3-\local]}-P_{[3-\psi^*]}}{P_{[3-\local]}}$; 
$P$ denotes the bound on perplexity (evaluated with $\psi^*$) 
and the subscript denotes the model and method used during training.
Each point on the x-axis is a restriction of the dataset to the top $L$
most frequently occurring words (number of features).
}
\label{fig:wiki_vs_features}
\end{figure}

\textbf{Sparse data is challenging: }
What is the relationship between data sparsity and how well inference networks work? 
We hold fixed the number of training samples and vary the sparsity of the data.
We do so by restricting the Wikipedia dataset to the top $L$ most frequently occurring words. 
We train three layer generative models on the different subsets. 
On training and held-out data, we computed the difference between 
the perplexity when the model is trained with (denoted $P_{[\Model{3}{\psi^*}]}$)
and without optimization of $\local$ (denoted $P_{[\Model{3}{\local}]}$). We plot the relative
decrease in perplexity obtained by training with $\psi^*$ in Fig. \ref{fig:wiki_vs_features}. 

Learning with $\psi^*$ helps \emph{more} as the data dimensionality
increases.  Data sparsity, therefore, poses a significant challenge to
inference networks.  One possible explanation is that many of the
tokens in the dataset are rare, and the inference network therefore
needs many sweeps over the dataset to learn to properly interpret
these rare words; while the inference network is learning to interpret
these rare words the generative model is receiving essentially random
learning signals that drive it to a poor local optimum.

Designing new strategies that can deal with such
data may be a fruitful direction for future work. This may require new 
architectures or algorithms---we found that simply making the inference network
deeper does not solve the problem. 

\begin{figure*}[h!]
\centering
\begin{subfigure}{.25\linewidth}
  \centering
  \includegraphics[width=\linewidth]{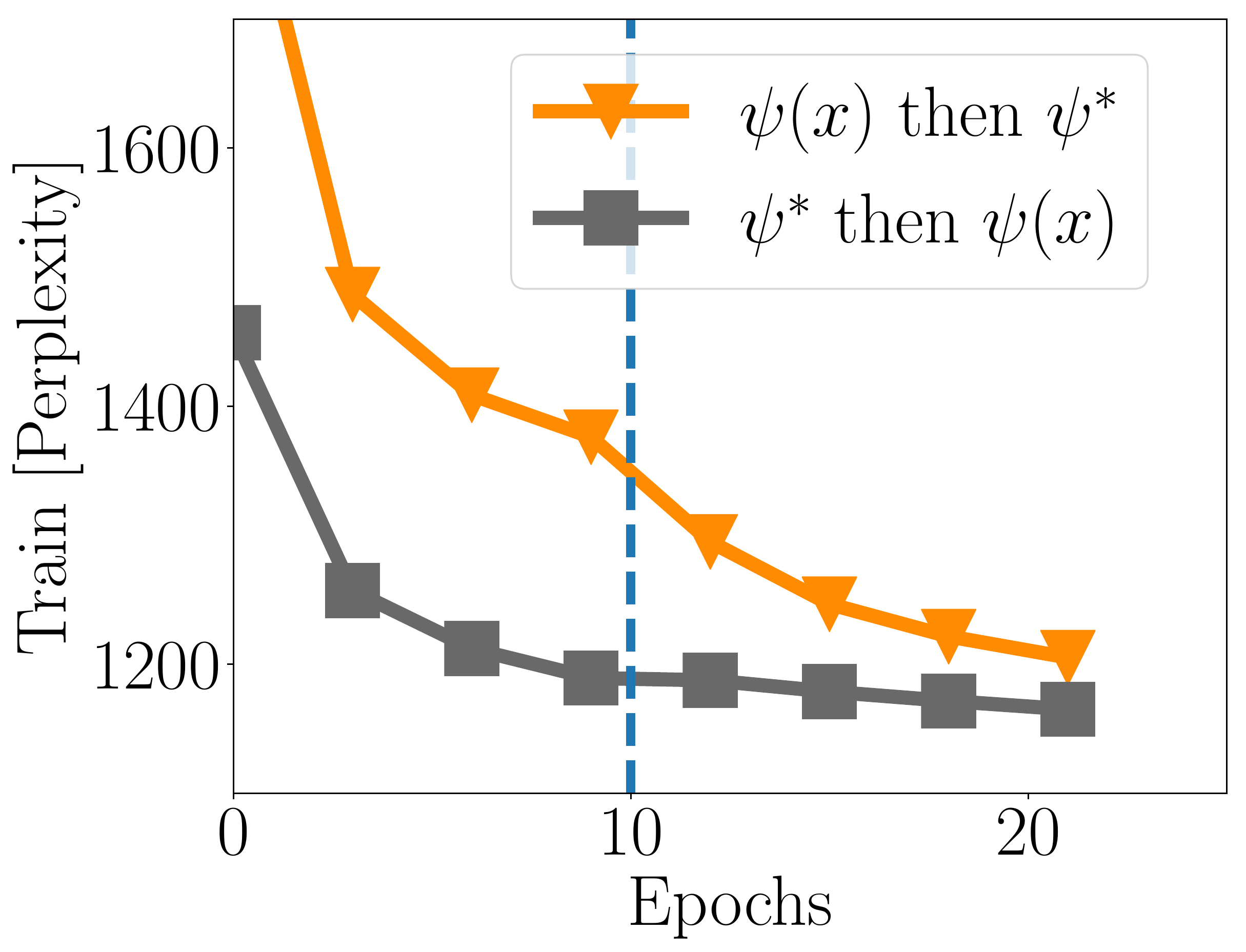}
  \caption{Training Data}
  \label{fig:wiki_train_bounds_mixed}
\end{subfigure}
\begin{subfigure}{.25\linewidth}
	\includegraphics[width=\linewidth]{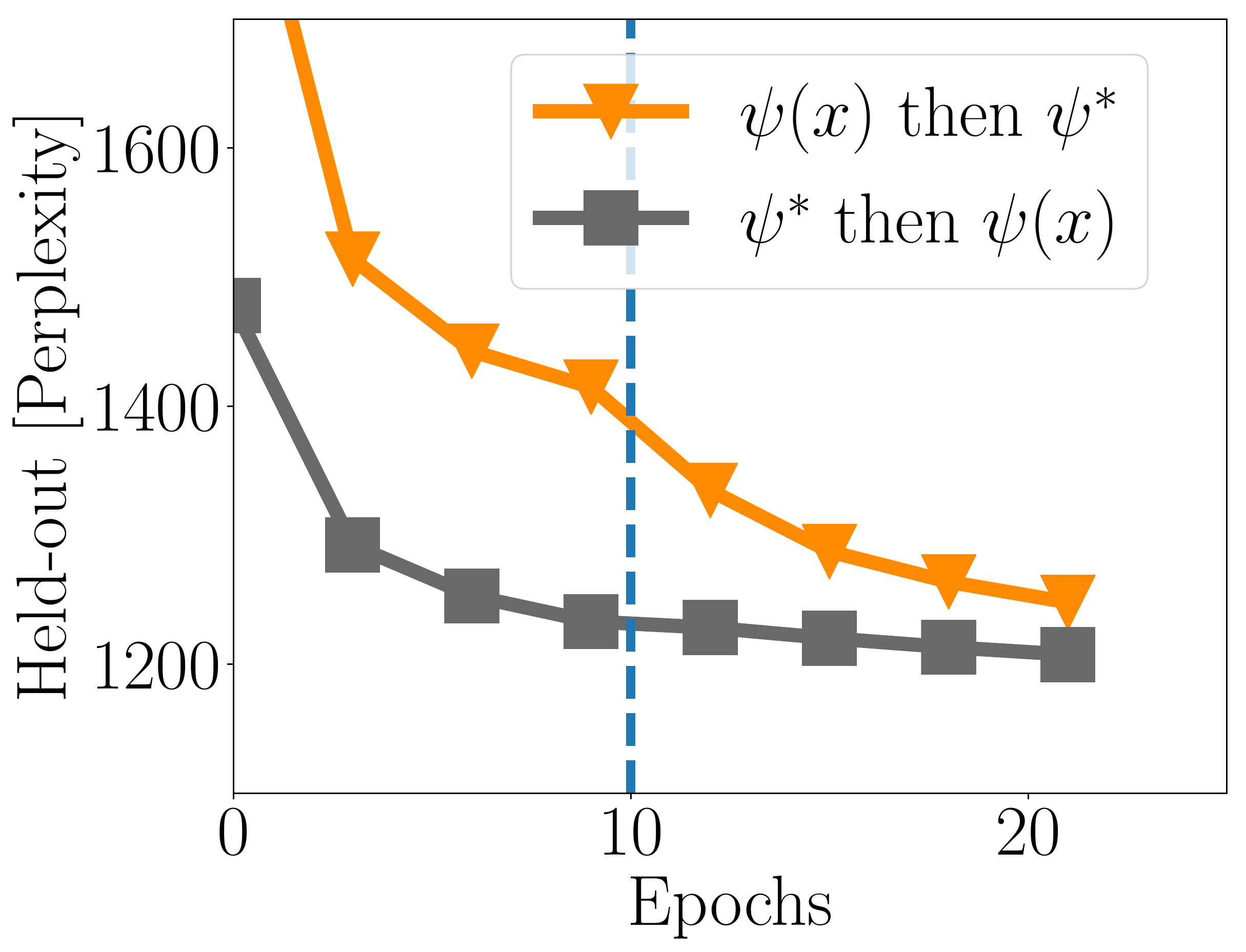}
	\caption{Held-out Data}
\label{fig:wiki_eval_bounds_mixed}
\end{subfigure}
\begin{subfigure}{.4\linewidth}
	\includegraphics[width=\linewidth]{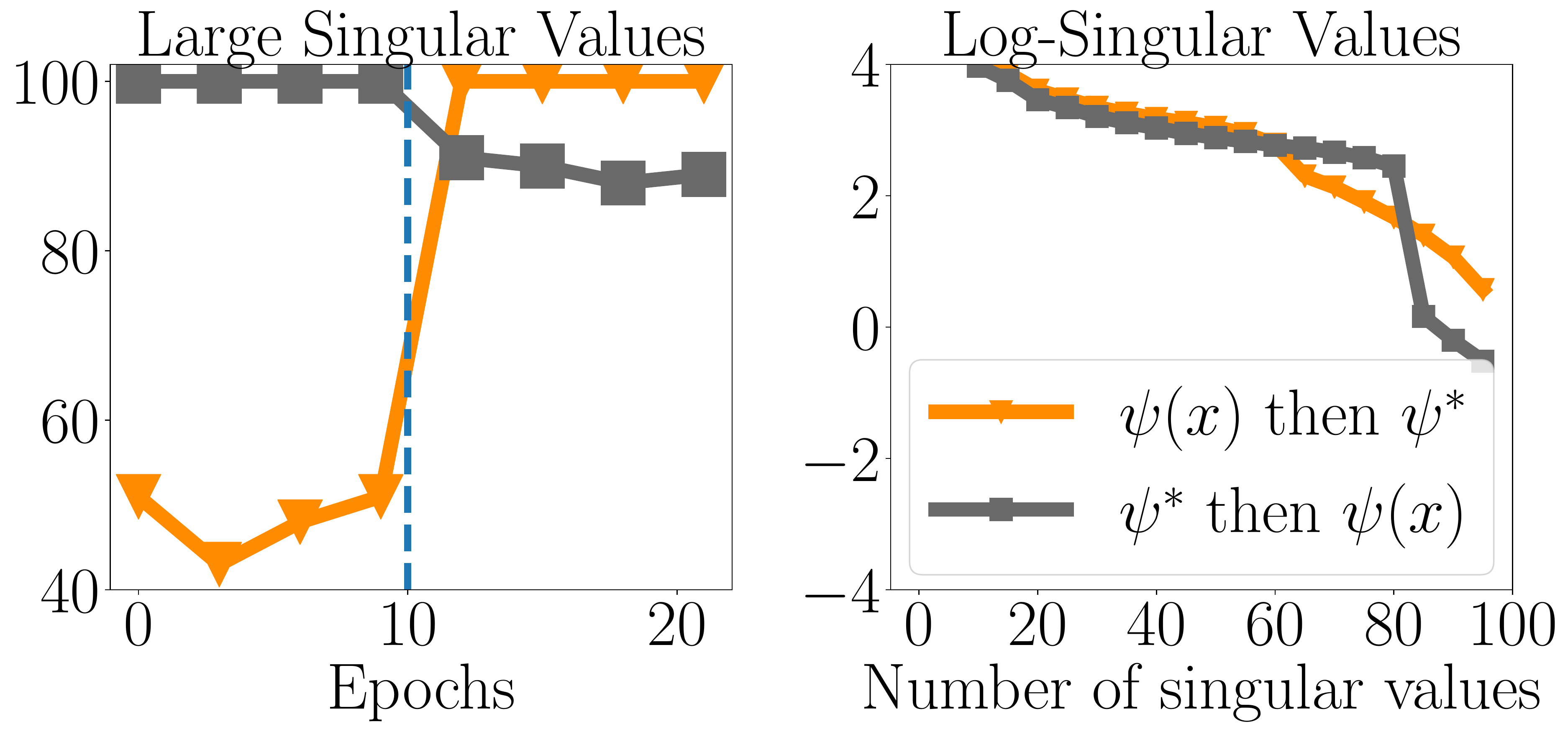}
\caption{Log-singular Values}
\label{fig:wiki_singular_mixed}
\end{subfigure}
\caption{
\small
\textbf{Late versus Early optimization of $\local$: }
Fig. \ref{fig:wiki_train_bounds_mixed} (\ref{fig:wiki_eval_bounds_mixed}) denote the train (held-out)
perplexity for three-layered models trained on the Wikipedia data in the following scenarios: 
$\psi^*$ is used for training for the first 
ten epochs following which $\local$ is used (denoted ``$\localopt$ then $\local$'') 
and vice versa (denoted ``$\local$ then $\localopt$'').
Fig. \ref{fig:wiki_singular_mixed} (Left) depicts the number 
of singular values of the Jacobian matrix $\nabla_z \log p(x|z)$ 
with value greater than $1$ as a function of training 
epochs for each of the two aforementioned methodologies.
Fig. \ref{fig:wiki_singular_mixed} (Right) plots the sorted log-singular
values of the Jacobian matrix corresponding to the final model under each training strategy. 
}
\label{fig:mixed_training}
\end{figure*}

\begin{figure}[h]
\centering
\includegraphics[width=0.7\linewidth]{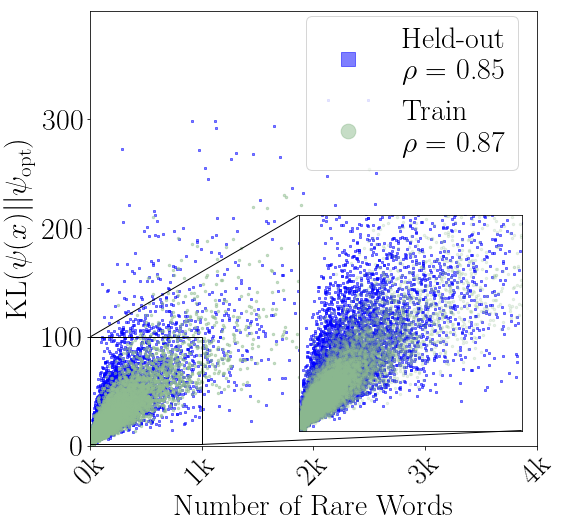}
\caption{
\small
\textbf{Raw KL and Rare Word Counts: }
We plot the raw values of $\text{KL}(\local \| \localopt)$
versus the number of rare words. We zoom into the plot and reduce the 
opacity of the train points to better see the held-out points. 
The Spearman $\rho$ correlation coefficient is computed 
between the two across $20,000$ points. We find a positive correlation. 
}
\label{fig:kl_vs_rare}
\end{figure}

\textbf{When should $\local$ be optimized: }
\emph{When} are the gains obtained from learning with $\psi^*$ accrued?
We learn three-layer models on Wikipedia under two settings: 
(a) we train for $10$ epochs using $\psi^*$ and then $10$ epochs using $\local$.
and (b) we do the opposite. 

Fig. \ref{fig:mixed_training} depicts the results of this experiment.
We find that: (1) much of the gain from optimizing $\local$ comes from the early epochs, (2) somewhat surprisingly
using $\psi^*$ instead of $\psi(x)$ later on in learning also helps (as witnessed by the sharp drop in perplexity after epoch $10$ and the number
of large singular values in Fig. \ref{fig:wiki_singular_mixed} [Left]). This suggests that even after seeing the data for several passes, the
inference network is unable to find $\local$ that explain the data well.
Finally, (3) for a fixed computational budget, one is better off
optimizing $\local$ sooner than later -- the curve that optimizes $\local$ later on does not catch up to the one
that optimizes $\local$ early in learning. 
This suggests that learning early with $\psi^*$, even for a few epochs, may alleviate underfitting. 

\textbf{Rare words and loose lower bounds: }
Fig. \ref{fig:wiki_vs_features} suggests that data sparsity presents a problem for inference networks at an aggregate level.  
We now ask which \emph{data points} benefit from the optimization of $\local$?
We sample $20000$ training and held-out data points; we compute $\text{KL}(\local \| \localopt)$ (both 
are Normal distributions and the KL
is analytic) and the number of rare words in each document (where a word is classified as being 
rare if it occurs in less than $5\%$ of training documents).
We visualize them in Fig. \ref{fig:kl_vs_rare}. We also display the Spearman $\rho$ correlation
between the two values in Fig. \ref{fig:kl_vs_rare}. There exists a positive 
correlation (about $0.88$ on the training data) between the two values suggesting that 
the gains in perplexity that we observe empirically in Table \ref{tab:perplexity} and Fig. \ref{fig:bounds} are due to
being able to better model the likelihood of documents with rare words in them. 

\textbf{Putting it all together: }
Our analysis describes a narrative of how underfitting 
occurs in learning VAEs on sparse data. The rare words in sparse, high-dimensional data are difficult to map
into local variational parameters that model the term $\mathbb{E}[\log p(x|z)]$ well (Fig. \ref{fig:wiki_vs_features},\ref{fig:kl_vs_rare}); 
$\qzx$ therefore focuses on the less noisy (the KL is evaluated analytically) signal of
minimizing $\KL(\qzx||p(z))$. Doing so prunes out many latent dimensions early on
resulting in underfitting (Fig. \ref{fig:wiki_singular_mixed} [Left]).
By using $\localopt$, the inadequacies of the inference
network are decoupled from the variational parameters used to derive gradients to $\theta$. 
The tighter variational bound $\optelbo$ achieves
a better tradeoff between $\mathbb{E}[\log p(x|z)]$ and $\KL(\qzx||p(z))$ (evidenced
by the number of large singular values of $\nabla_z \log p(x|z)$ when optimizing $\psi^*$ in Fig. \ref{fig:wiki_singular_mixed}).
The gradient updates with respect to this tighter bound better utilize $\theta$.

\subsection{Collaborative filtering\label{sec:recsys}}
Modeling rare features in sparse, high-dimensional data is necessary to achieve strong results on this task.  
We study the top-N recommendation performance of NFA under strong 
generalization \citep{marlin2009collaborative}. 

\textbf{Datasets: }
We study two large user-item rating datasets: MovieLens-20M (ML-20M) 
\citep{harper2015movielens} and Netflix\footnote{\url{http://www.netflixprize.com/}}. 
Following standard procedure: we binarize the explicit rating data, keeping ratings of four or higher and interpreting 
them as implicit feedback \citep{hu2008collaborative} and keep users who have positively rated at least five movies.
We train with users' binary implicit feedback as $x_{d}$; the vocabulary is the set of all movies. 
The number of training/validation/test users is 116,677/10,000/10,000 for ML-20M ($V$: 20,108) 
and 383,435/40,000/40,000 for Netflix ($V$: 17,769).

\begin{table}[b!]
\centering
\caption{
\small
\textbf{Recall and NDCG on Recommender Systems:} 
``\Model{2}{$\psi^*$}-tfidf'' denotes a two-layer 
(one hidden layer and one output layer) generative model.
Standard errors are around 0.002 for ML-20M and 0.001 for Netflix.  
{\bf Runtime:} WMF takes on the order of minutes [ML-20M \& Netflix]; CDAE and NFA ($\local$) take ~8 hours [ML-20M]
and ~32.5 hours [Netflix] for $150$ epochs; NFA ($\localopt$) takes takes 1.5 days [ML-20M] and 3 days [Netflix]; 
SLIM takes 3-4 days [ML-20M] and 2 weeks [Netflix].
}
\label{tab:sg_results}
\small
\begin{subtable}[t]{\columnwidth}
	\centering
\begin{tabular}{ c c c c c}
      \multicolumn{1}{c}{\textbf{ML-20M}}     & \multicolumn{2}{c}{Recall@50} 
    & \multicolumn{2}{c}{NDCG@100}   \\   \toprule
     {\bf NFA} & $\local$ & $\localopt$ & $\local$ &$\localopt$\\
  \Model{2}{$\local$}-norm  & 0.475 & 0.484 & 0.371 & 0.377 \\ 
  \Model{2}{$\localopt$}-norm & 0.483 & 0.508 & 0.376 & 0.396 \\
  \Model{2}{$\local$}-tfidf  & 0.499 & 0.505 & 0.389 & 0.396 \\
  \Model{2}{$\localopt$}-tfidf & 0.509 & \textbf{0.515} & 0.395 & \textbf{0.404}\\ \midrule
    \sc wmf & \multicolumn{2}{c}{0.498} & \multicolumn{2}{c}{0.386}  \\ 
    \sc slim & \multicolumn{2}{c}{0.495} & \multicolumn{2}{c}{0.401}\\
    \sc cdae & \multicolumn{2}{c}{0.512} & \multicolumn{2}{c}{0.402}\\
\bottomrule
\end{tabular}
\end{subtable}
\\
\vspace{0.1in}
\begin{subtable}[t]{\columnwidth}
\centering
\begin{tabular}{ c c c c c}
   \multicolumn{1}{c}{\textbf{Netflix}}     & \multicolumn{2}{c}{Recall@50} & \multicolumn{2}{c}{NDCG@100}   \\   \toprule
     \bf{NFA}& $\local$ & $\localopt$ & $\local$ &$\localopt$\\
  \Model{2}{$\local$}-norm   & 0.388 & 0.393 & 0.333 & 0.337        \\ 
  \Model{2}{$\localopt$}-norm  & 0.404 & 0.415 & 0.347 & 0.358\\
  \Model{2}{$\local$}-tfidf  & 0.404 & 0.409 & 0.348 & 0.353\\
  \Model{2}{$\localopt$}-tfidf & 0.417 & 0.424 & 0.359 & 0.367 \\\midrule
    \sc wmf & \multicolumn{2}{c}{0.404} & \multicolumn{2}{c}{0.351}  \\ 
    \sc slim & \multicolumn{2}{c}{\textbf{0.427}} & \multicolumn{2}{c}{\textbf{0.378}}\\
    \sc cdae & \multicolumn{2}{c}{0.417} & \multicolumn{2}{c}{0.360}\\
\bottomrule
\end{tabular}
\end{subtable}

\end{table}

\textbf{Evaluation and Metrics: } 
We train with the complete feedback history from training users, and evaluate on held-out validation/test users. We select model architecture (MLP with 0, 1, 2 hidden layers) from the held-out validation users based on NDCG@100 and report metrics on the held-out test users.  
For held-out users, we randomly select $80\%$ of the feedback as the input to the inference network and see how the other $20\%$ of the positively rated items are ranked based 
$\mu(z)$. 
We report two ranking-based metrics averaged over all held-out users: Recall$@N$ and 
truncated normalized discounted cumulative gain (NDCG$@N$) \citep{jarvelin2002cumulated}. For each user,
both metrics compare the predicted rank of unobserved items with their true
rank. While Recall$@N$ considers all items ranked within the first $N$ to be equivalent, NDCG$@N$ uses a monotonically increasing 
discount to emphasize the importance of higher ranks versus lower ones. 

Define $\pi$ as a ranking over all the items where $\pi(v)$ indicates the $v$-th 
ranked item, $\mathbb{I}\{\cdot\}$ is the indicator function, and $d(\pi(v))$ returns $1$ if user $d$ has positively rated item
$\pi(v)$. Recall@$N$ for user $d$ is
\[
  \textrm{Recall}@N(d,\pi) := \sum_{v=1}^N \frac{\mathbb{I}\{d(\pi(v))=1\}}{\text{min}(N, \sum_{v'}^V \mathbb{I}\{d(\pi(v'))=1\})}.
\]
The expression in the denominator evaluates to the minimum between $N$ and the
number of items consumed by user $d$. This normalizes Recall@$N$ to have a maximum of 1, which corresponds to ranking all relevant items in the top $N$ positions. 
Discounted cumulative gain (DCG@$N$) for user $d$ is
\[
  \text{DCG}@N(d,\pi) := \sum_{v=1}^N \frac{ 2^{\mathbb{I}\{d(\pi(v))=1\}} -1}{ \log(v + 1) }.
\]
NDCG@$N$ is the DCG$@N$ normalized by ideal DCG\@$N$, where all the relevant items are ranked at the top. We have, NDCG@$N$ $\in [0, 1]$.
As baselines, we consider:

\textbf{Weighted matrix factorization (WMF)} \citep{hu2008collaborative}: a linear low-rank factor model. 
We train WMF with alternating least squares; this generally leads to better performance than with SGD.\\ 
\textbf{SLIM} \citep{ning2011slim}: a linear model which learns a sparse item-to-item similarity 
matrix by solving a constrained $\ell_1$-regularized optimization problem. \\
\textbf{Collaborative denoising autoencoder (CDAE)} \citep{wu2016collaborative}: An autoencoder achitecture 
specifically designed for top-N recommendation. It augments a denoising autoencoder 
\citep{vincent2008extracting} by adding a per-user latent vector to the input,
inspired by standard linear matrix-factorization approaches. Among the baselines, CDAE is most akin to NFA. 

Table~\ref{tab:sg_results} summarizes the results of NFA under different settings. 
We found that optimizing $\local$ helps both at train and test time and that TF-IDF features consistently improve performance.
Crucially, the standard training procedure for VAEs realizes a poorly trained model that underperforms every baseline.
The improved training techniques we recommend generalize across different kinds of sparse data.
With them, \emph{the same generative model}, outperforms CDAE and WMF on both datasets, 
and marginally outperforms SLIM on
ML-20M while achieving nearly state-of-the-art results on Netflix. In terms of runtimes, we found that learning
NFA (with $\psi^*$) to be approximately two-three times faster than SLIM.
Our results highlight the importance of inference at training time showing 
NFA, when properly fit, can outperform the popular linear factorization approaches.

\section{Discussion\label{sec:discussion}}
Studying the failures of learning with inference networks is an important step to designing more robust 
neural architectures for inference networks. 
We show that avoiding gradients obtained using poor variational parameters
is vital to successfully learning VAEs on sparse data.
An interesting question is \emph{why} inference networks have a harder time 
turning sparse data into variational parameters compared to images?
One hypothesis is that the redundant correlations
that exist among pixels (but occur less frequently in features found in
sparse data) are more easily transformed into local variational parameters $\local$ that are, 
in practice, often reasonably close to $\localopt$ during learning. 

\section*{Acknowledgements}
The authors are grateful to David Sontag, Fredrik Johansson, 
Matthew Johnson, and Ardavan Saeedi for helpful
comments and suggestions regarding this work.

\bibliographystyle{authordate1}
\bibliography{refs}

\twocolumn[

	\aistatstitle{Supplementary Material}

]
\section*{Contents}

\begin{enumerate}
	\item Spectral analysis of the Jacobian matrix
	\item Relationship with annealing the KL divergence
	\item Inference on documents with rare words
	\item Depth of $\qzx$
	\item Learning with $\psi^*$ on small dataset
\end{enumerate}

\section{Spectral Analysis of the Jacobian Matrix}
For any vector valued function $f(x): \mathbb{R}^K\rightarrow \mathbb{R}^V$, 
$\nabla_x f(x)$ is the matrix-valued function representing the sensitivity of the output to the input. 
When $f(x)$ is a deep neural network, \cite{wang2016analysis} use the spectra of the Jacobian matrix under various inputs $x$ 
to quantify the complexity of the learned function. They find that the spectra are correlated with the complexity of the learned function.

We adopt their technique for studying the utilization of the latent space in deep generative models. 
In the case of NFA, we seek to quantify the learned complexity of the generative model.
To do so, we compute the Jacobian matrix 
as $\mathcal{J}(z)  = \nabla_z \log p(x|z)$. This is a read-out measure of the sensitivity of the likelihood
with respect to the latent dimension. 

$\mathcal{J}(z)$ is a matrix valued function that can be evaluated at every point in the latent space. We evaluate it 
it at the mode of the (unimodal) prior distribution i.e. at $z= \vec{0}$. The singular values of the resulting matrix
denote how much the log-likelihood changes from the origin along the singular vectors lying in latent space. The intensity
of these singular values (which we plot) is a read-out measure of 
how many intrinsic dimensions are utilized by the model parameters $\theta$ at the mode of the prior distribution.
Our choice of evaluating $\mathcal{J}(z)$ at 
$z=\vec{0}$ is motivated by the fact that much of the probability mass in latent space under the NFA
model will be placed at the origin. We use the utilization at the mode as an approximation for the utilization across the entire
latent space. 
We also plotted the spectral decomposition obtained under a Monte-Carlo approximation to the matrix
$\mathbb{E}[\mathcal{J}(z)]$ and found it to be similar to the 
decomposition obtained by evaluating the Jacobian at the mode. 

Another possibility to measure utilization would be from 
the KL divergence of the prior and the output 
of the inference network (as in \citet{burda2015importance}).

\section{Learning with $\psi^*$ on a small dataset}

\begin{table}[h!]
\caption{\small \textbf{Test Perplexity on 20newsgroups: } \textbf{Left: Baselines} 
Legend: LDA \citep{blei2003latent}, Replicated Softmax (RSM) \citep{hinton2009replicated}, 
Sigmoid Belief Networks (SBN) and Deep Autoregressive 
Networks (DARN) \citep{mnih2014neural}, 
Neural Variational Document Model (NVDM) \citep{miao2015neural}.
$K$ denotes the latent dimension in our notation.
\textbf{Right: } NFA on text data with $K=100$.  We vary the features presented
to the inference network $\qzx$ during learning between: 
normalized count vectors ($\frac{x}{\sum_{i=1}^{V} x_i}$, denoted ``norm'') 
and normalized TF-IDF (denoted ``tfidf'') features.}
	\label{tab:perplexity_20newsgroups}
	\resizebox{\linewidth}{!}{
	\centering
		\begin{tabular}{cccc}
			\multicolumn{1}{c}{\bf Model}  &\multicolumn{1}{c}{\bf $K$} &\multicolumn{1}{c}{\bf Results}\\
			\hline 
			LDA &50 &1091 \\
			LDA &200 &1058 \\
			RSM &50 &953 \\
			\hline
			SBN &50 &909 \\
			fDARN &50 &917 \\
			fDARN &200 &--- \\
			\hline
			NVDM &50 &836 \\
			NVDM &200 &852 \\
		\end{tabular}
		\quad    
		\begin{tabular}{c|cc}
      \multirow{2}{*}{\bf NFA}  &
      \multicolumn{2}{c}{\bf Perplexity}\\
      & $\local$ & $\localopt$\\
      \Model{1}{$\local$}-norm & 1018 & 903\\
      \Model{1}{$\localopt$}-norm  &1279 & 889\\
      \Model{3}{$\local$}-norm  & 986 & 857\\
      \Model{3}{$\localopt$}-norm &1292 & 879\\
      \hline
      \Model{1}{$\local$}-tfidf  & 932& 839 \\
      \Model{1}{$\localopt$}-tfidf & 953& 828\\
      \Model{3}{$\local$}-tfidf & 999& 842\\
      \Model{3}{$\localopt$}-tfidf & 1063 & 839\\
    \end{tabular}
}
\end{table}
\begin{figure*}[h!]
\centering
\begin{subfigure}{.3\linewidth}
  \centering
  \includegraphics[width=\linewidth]{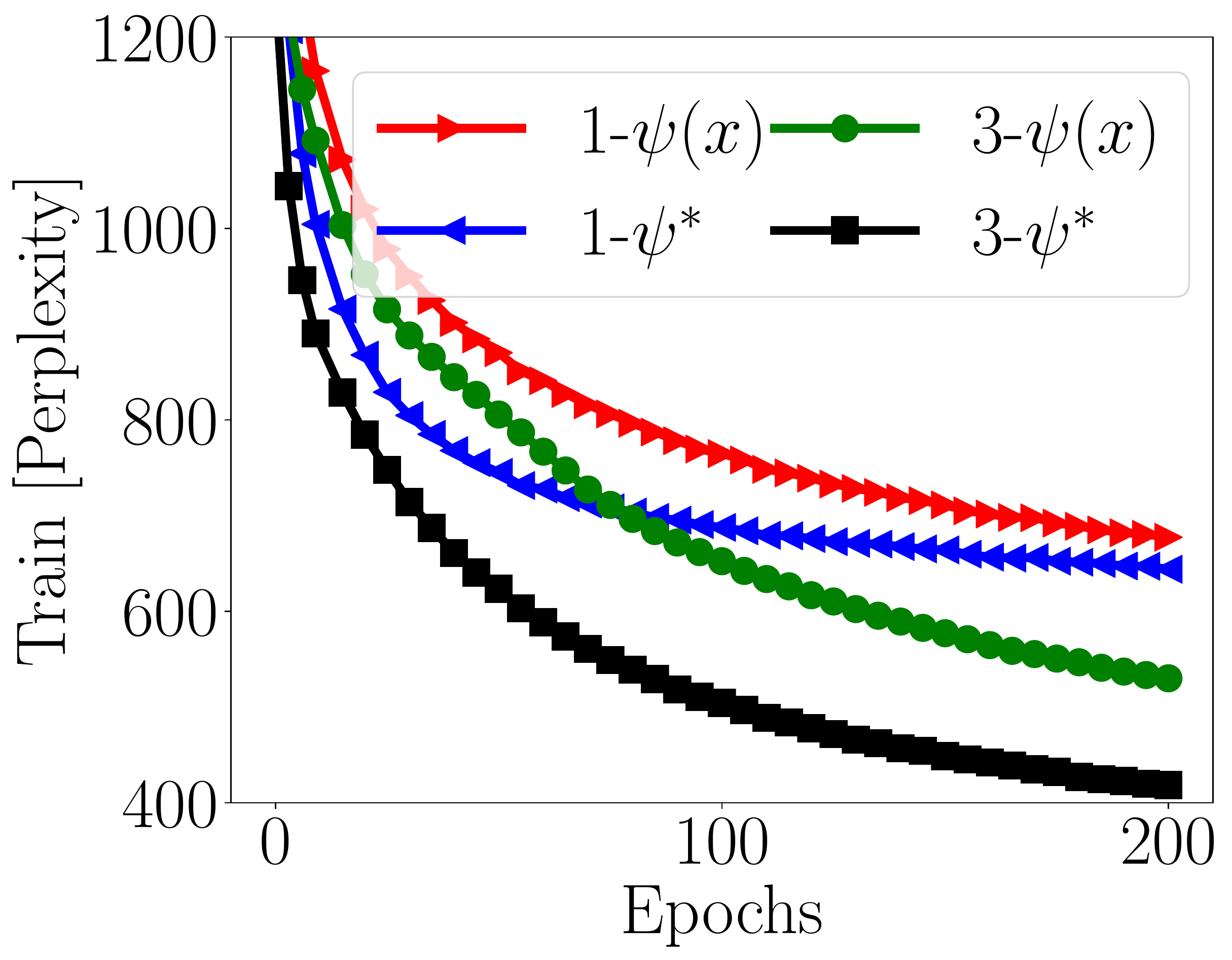}
  \caption{Training Data}
  \label{fig:news_train_bounds}
\end{subfigure}
\begin{subfigure}{.3\linewidth}
	\includegraphics[width=\linewidth]{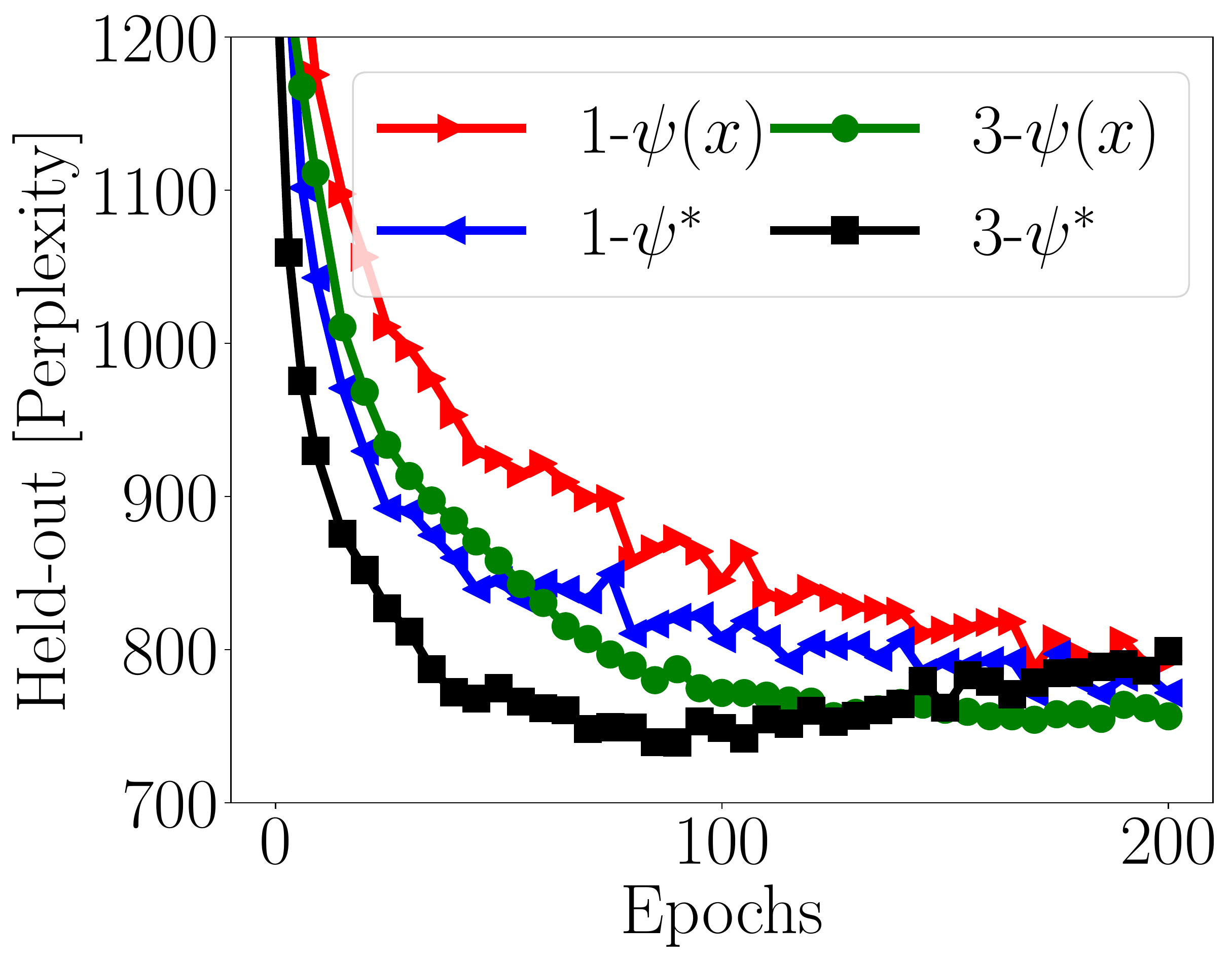}
	\caption{Held-out Data}
\label{fig:news_eval_bounds}
\end{subfigure}
\begin{subfigure}{.3\linewidth}
	\includegraphics[width=\linewidth]{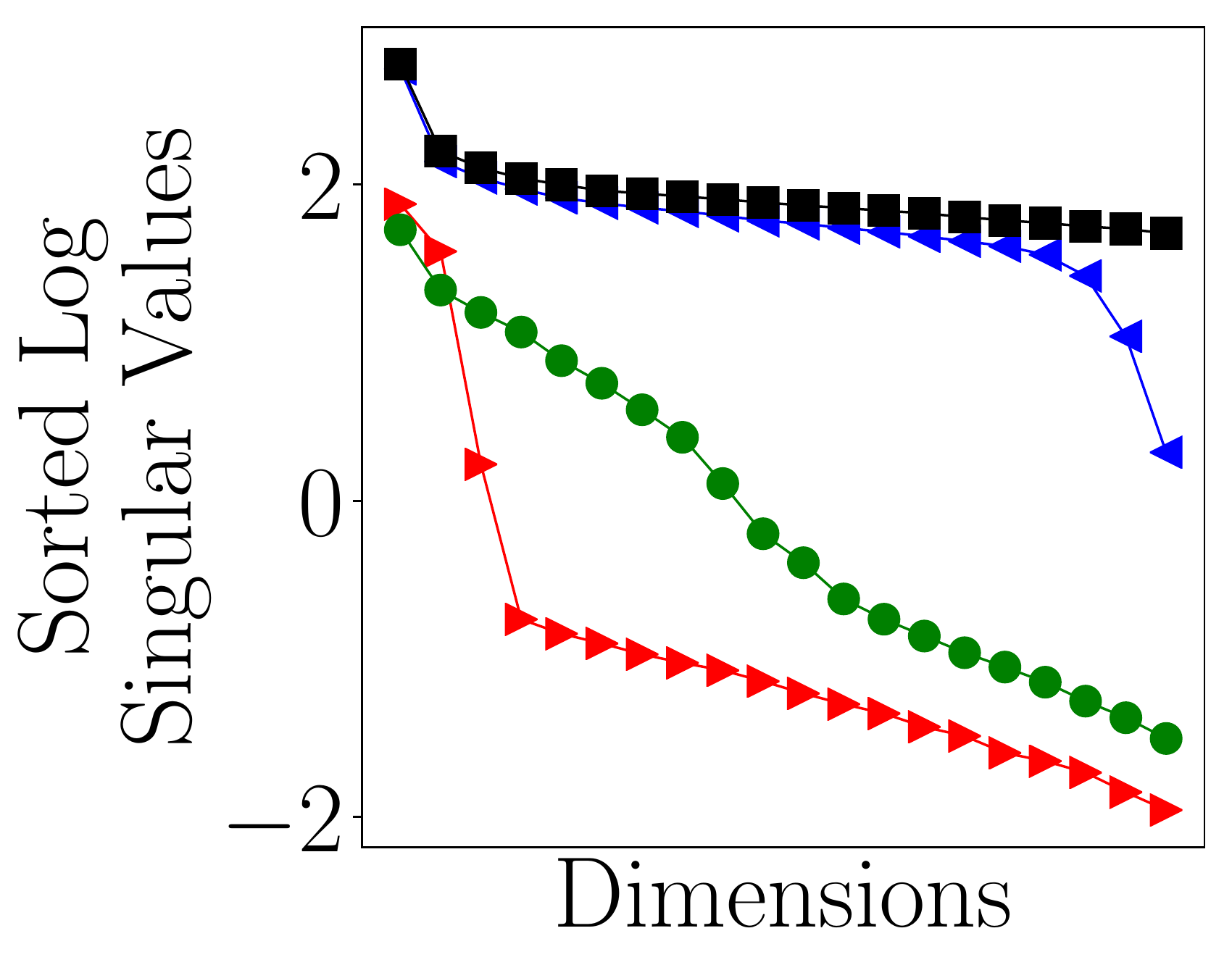}
\caption{Log-singular Values}
\label{fig:news_singular}
\end{subfigure}
\caption{
\small
\textbf{20Newsgroups - Training and Held-out Bounds: }
Fig. \ref{fig:news_train_bounds}, \ref{fig:news_eval_bounds} denotes the train (held-out)
perplexity for different models. Fig. \ref{fig:news_singular} depicts the log-singular values of the Jacobian matrix for the trained models. 
}
\label{fig:news_training}
\end{figure*}
\begin{figure*}[h!]
\centering
\begin{subfigure}{.3\linewidth}
  \centering
  \includegraphics[width=\linewidth]{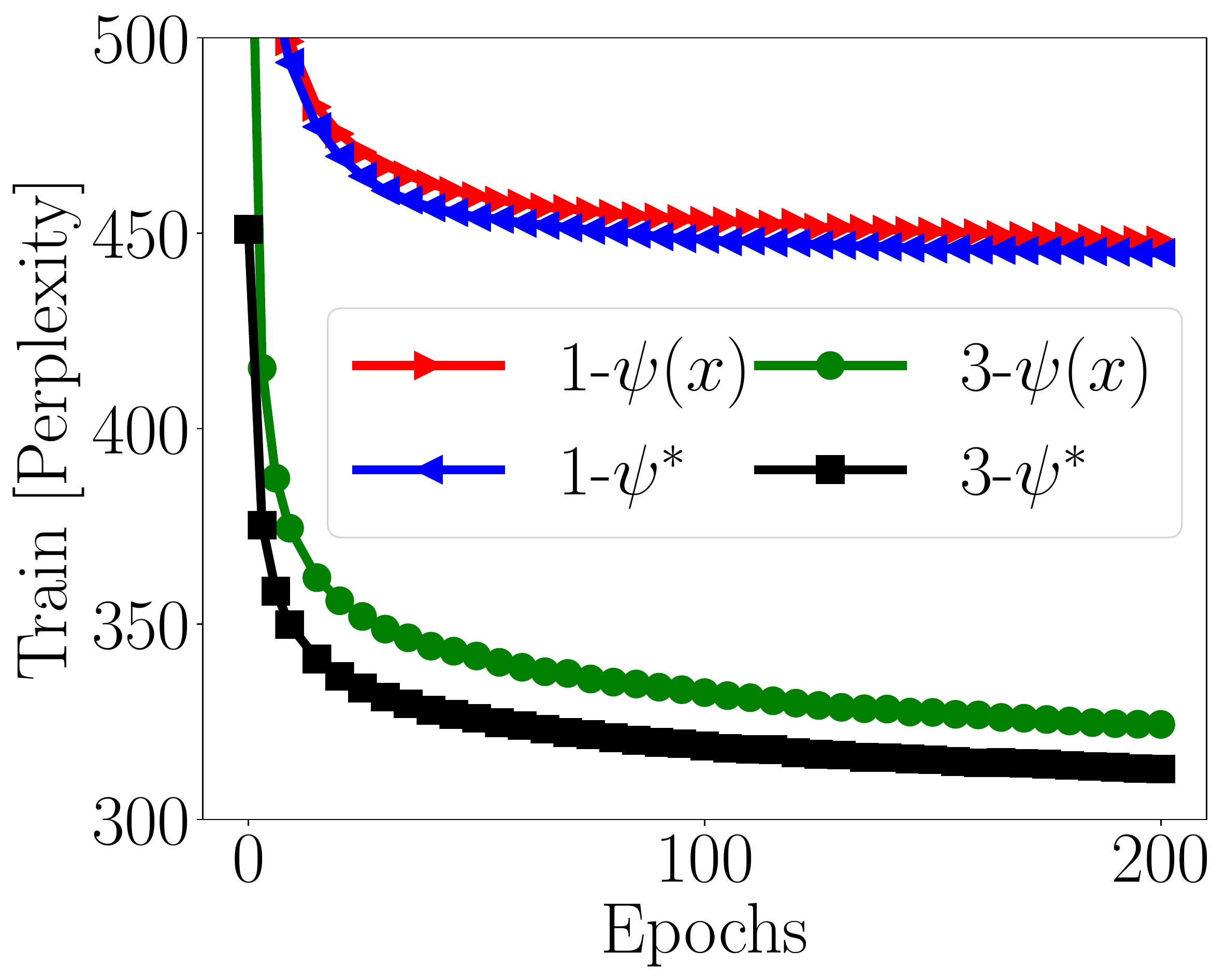}
  \caption{Training Data}
  \label{fig:rcv2_train_bounds}
\end{subfigure}
\begin{subfigure}{.3\linewidth}
	\includegraphics[width=\linewidth]{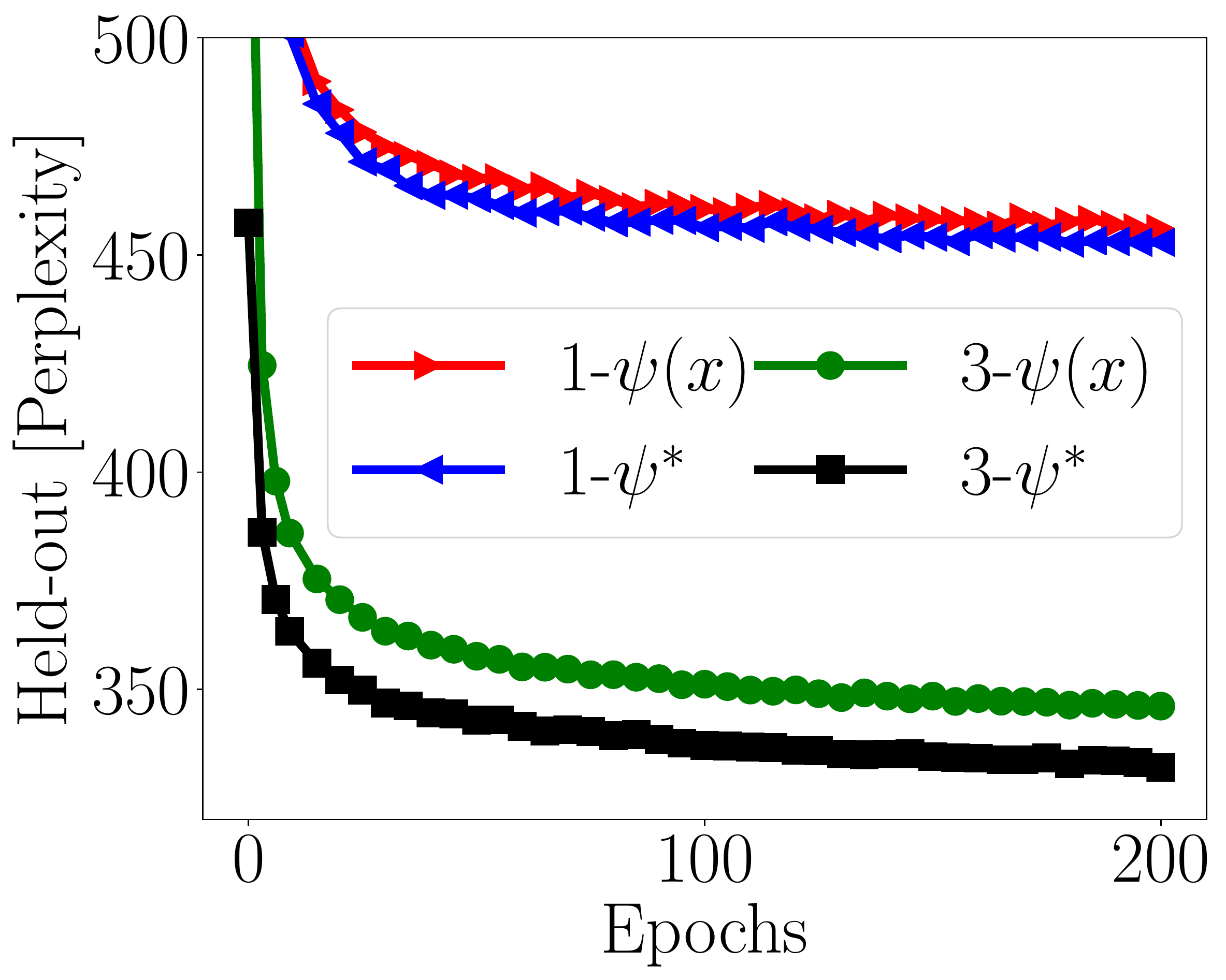}
	\caption{Held-out Data}
\label{fig:rcv2_eval_bounds}
\end{subfigure}
\begin{subfigure}{.3\linewidth}
	\includegraphics[width=\linewidth]{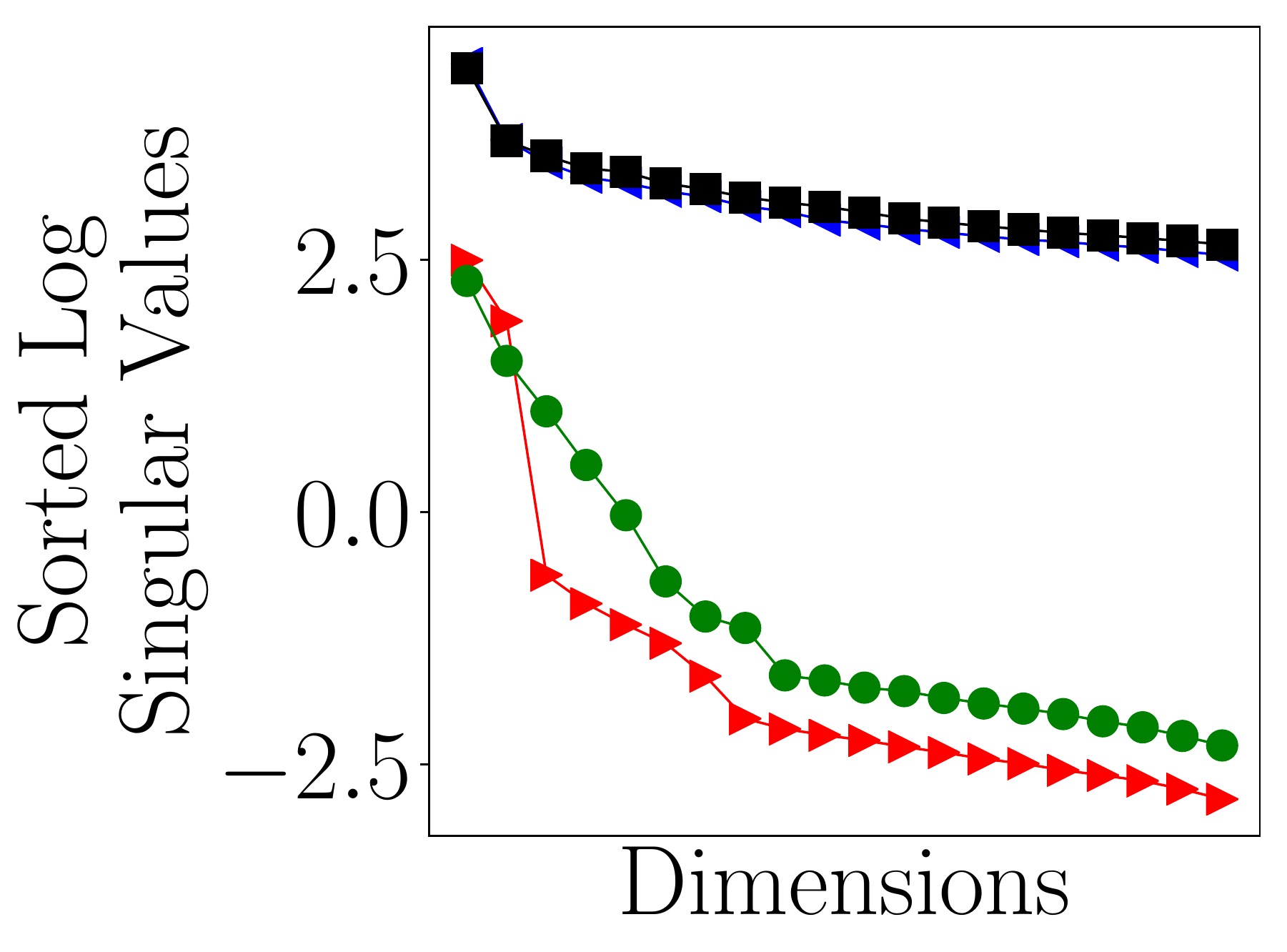}
\caption{Log-singular Values}
\label{fig:rcv2_singular}
\end{subfigure}
\caption{
\small
\textbf{RCV1 - Training and Held-out Bounds: }
Fig. \ref{fig:rcv2_train_bounds}, \ref{fig:rcv2_eval_bounds} denotes the train (held-out)
perplexity for different models. Fig. \ref{fig:rcv2_singular} depicts the 
log-singular values of the Jacobian matrix for the trained models. 
}
\label{fig:rcv2_training}
\end{figure*}

In the main paper, we studied the optimization of variational parameters on the larger RCV1 and Wikipedia datasets.
Here, we study the role of learning with $\localopt$ in the small-data regime. Table \ref{tab:perplexity_20newsgroups}
depicts the results obtained after training models for $200$ passes through the data. We summarize our findings: 
(1) across the board, TF-IDF features improve learning, and 
(2) in the small data regime, deeper non-linear models (\Model{3}{$\psi^*$}-tfidf) overfit 
quickly and better results are obtained by the simpler multinomial-logistic PCA model (\Model{1}{$\psi^*$}-tfidf).
Overfitting is also evident in Fig. \ref{fig:news_training} from 
comparing curves on the validation set to those on the training set. Interestingly, in the small dataset setting, 
we see that learning with $\local$ has the potential to have a regularization effect in that the results obtained
are not much worse than those obtained from learning with $\localopt$. 

For completeness, in Fig. \ref{fig:rcv2_training}, we also provide the 
training behavior for the RCV1 dataset corresponding to the results of Table \ref{tab:perplexity}. The results here, echo the convergence behavior on the Wikipedia dataset.

\section{Comparison with KL-annealing}

An empirical observation made in previous work is that when $p(x|z;\theta)$ is complex (parameterized by a recurrent neural 
network or a neural autoregressive density estimator (NADE)), the generative model also must contend with 
overpruning of the latent dimension. 
A proposed fix is the annealing of the KL divergence term 
in Equation \ref{eqn:varlowbnd} \citep[e.g.,][]{bowman2015generating} as one way to overcome local minima.
As discussed in the main paper, this is a different failure mode to the one we present in that 
our decoder is a vanilla MLP -- nonetheless, we apply KL annealing within our setting.

In particular, we optimized $\Exp{\log \pxz)}{\qzx} - \eta \KL(\, \qzx || p(z)\, )$ where $\eta$ was annealed from $0$
to $1$ (linearly -- though we also tried exponential annealing)
over the course of several parameter updates. Note that doing so does \emph{not} give us a lower bound on the likelihood of the data 
anymore. There are few established guidelines about the rate of annealing the KL divergence and in general, we found
it tricky to get it to work reliably. We experimented with different rates of annealing for learning a three-layer generative
model on the Wikipedia data. 

\begin{figure*}[t!]
\centering
\begin{subfigure}{.3\linewidth}
  \centering
  \includegraphics[width=\linewidth]{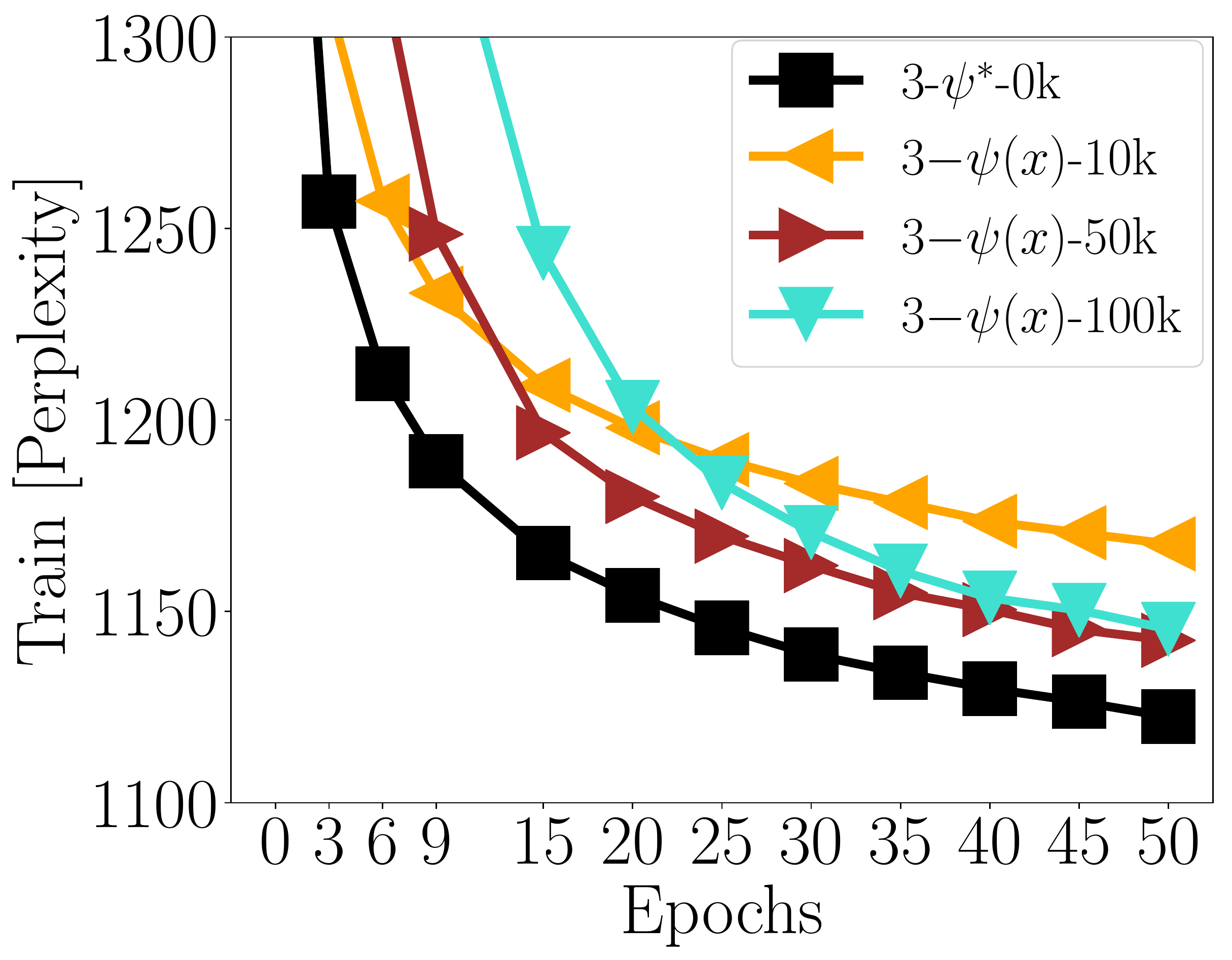}
  \caption{Training Data}
  \label{fig:anneal_train_bounds}
\end{subfigure}
\begin{subfigure}{.3\linewidth}
	\includegraphics[width=\linewidth]{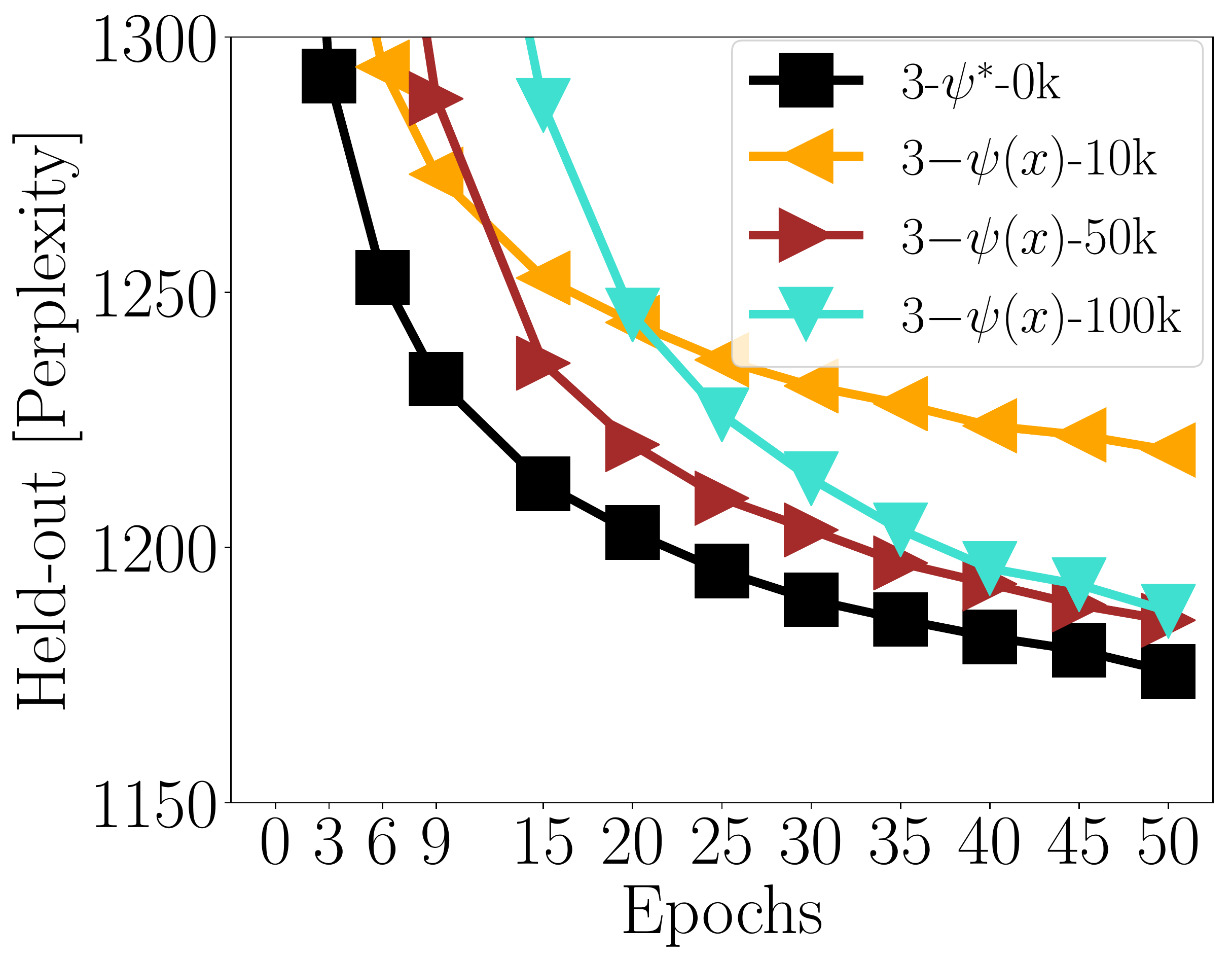}
	\caption{Held-out Data}
\label{fig:anneal_eval_bounds}
\end{subfigure}
\begin{subfigure}{.3\linewidth}
	\includegraphics[width=\linewidth]{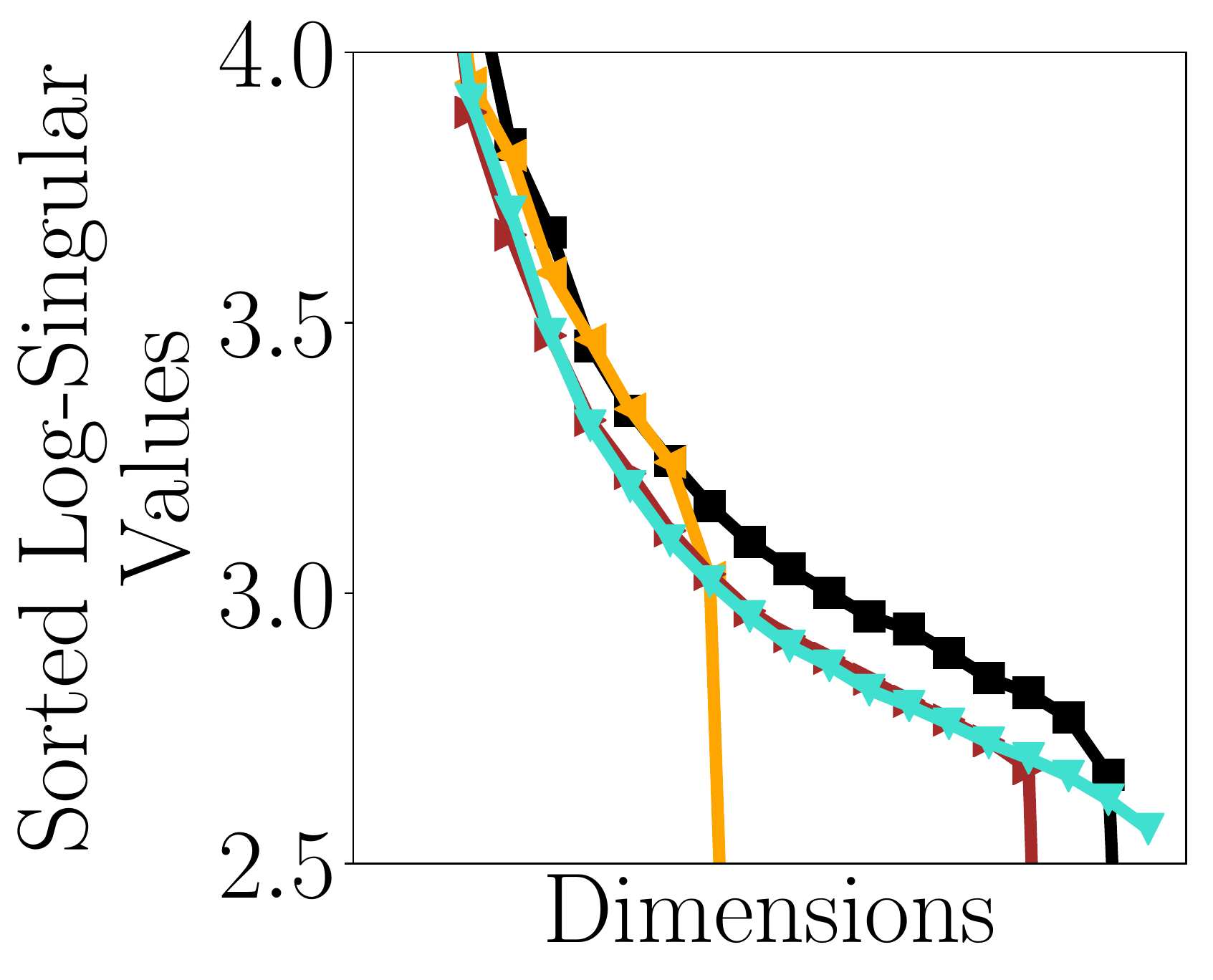}
\caption{Log-singular Values}
\label{fig:anneal_singular}
\end{subfigure}
\caption{
\small
\textbf{KL annealing vs learning with $\localopt$}
Fig. \ref{fig:anneal_train_bounds}, \ref{fig:anneal_eval_bounds} denotes the train (held-out)
perplexity for different training methods. The suffix at the end of the model configuration denotes
the number of parameter updates that it took for the KL divergence in Equation \ref{eqn:varlowbnd} to be annealed
from $0$ to $1$. \Model{3}{$\psi^*$}-50k denotes that it took $50000$ parameter updates before $-\elbo$ was used as the 
loss function. 
Fig. \ref{fig:news_singular} depicts the log-singular values of the Jacobian matrix for the trained models.
}
\label{fig:anneal_curves}
\end{figure*}

Our findings (visualized in Fig. \ref{fig:anneal_curves}) are 
as follows: (1) on sparse data we found annealing the KL divergence is 
very sensitive to the annealing rate -- too small
an annealing rate and we were still left with underfitting (as in annealing for 10k), too high an annealing rate (as in 
100k) and this resulted in slow convergence; 
(2) learning with $\psi^*$ always outperformed (in both rate of convergence and quality of final result on train and held-out data)
annealing the KL divergence across various choices of annealing schedules. 
Said differently, on the Wikipedia dataset, we conjecture 
there exists a choice of annealing of the KL divergence for which the perplexity obtained \emph{may}
match those of learning with $\psi^*$ but finding this schedule requires 
significant trial and error -- Fig. \ref{fig:anneal_curves} 
suggests that we did not find it. We found that learning with $\localopt$ was more robust, required
less tuning (setting values of $M$ to be larger than $100$ never hurt) and always performed at par or better than annealing
the KL divergence. Furthermore, we did not find annealing the KL to work effectively for the experiments on the recommender systems
task. In particular, we were unable to find an annealing schedule that reliably produced good results.

\section{Depth of $\qzx$}

\begin{figure*}[t!]
\centering
\begin{subfigure}{.3\linewidth}
  \centering
  \includegraphics[width=\linewidth]{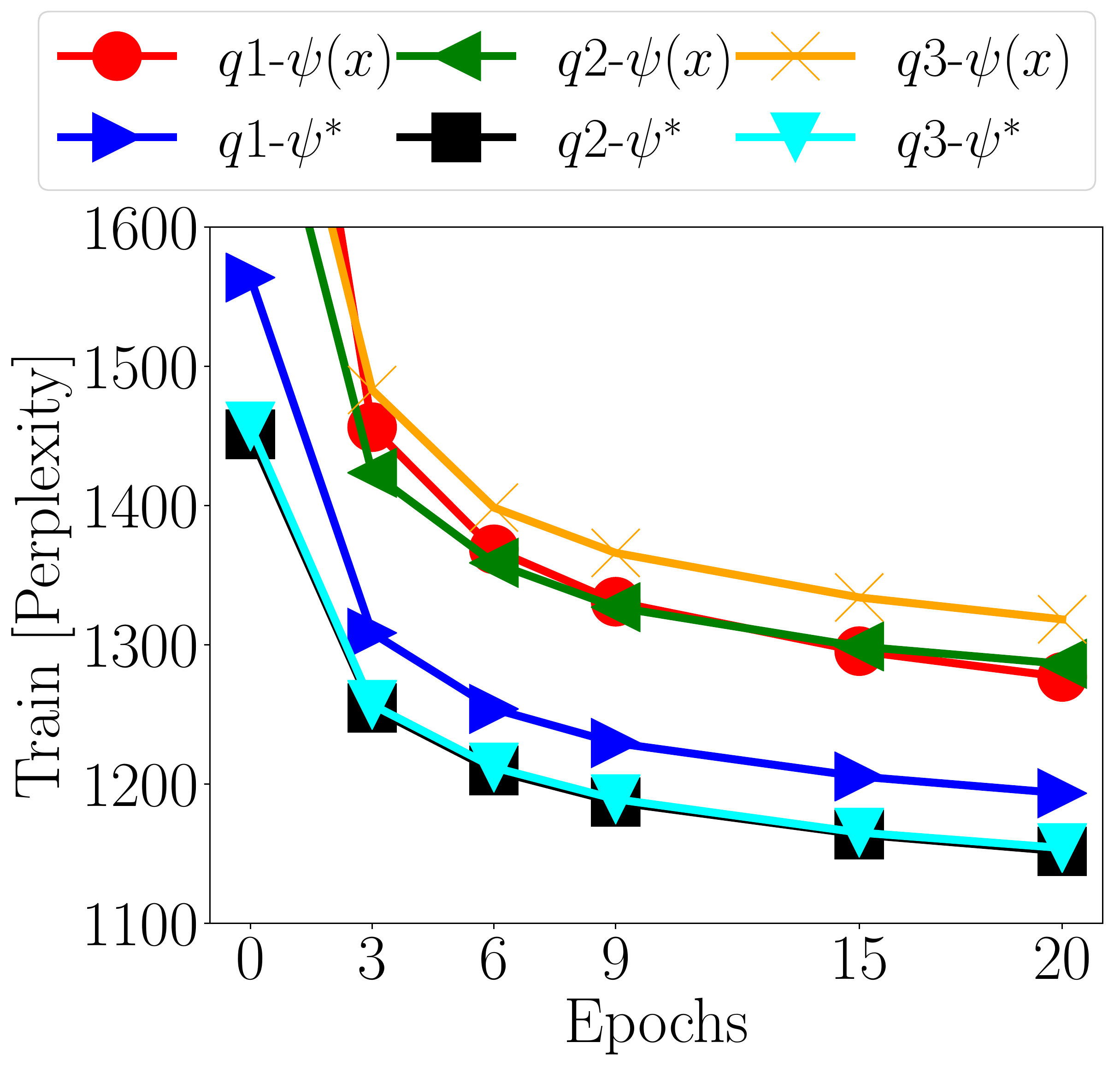}
  \caption{Training Data}
  \label{fig:qvary_train_bounds}
\end{subfigure}
\begin{subfigure}{.3\linewidth}
	\includegraphics[width=\linewidth]{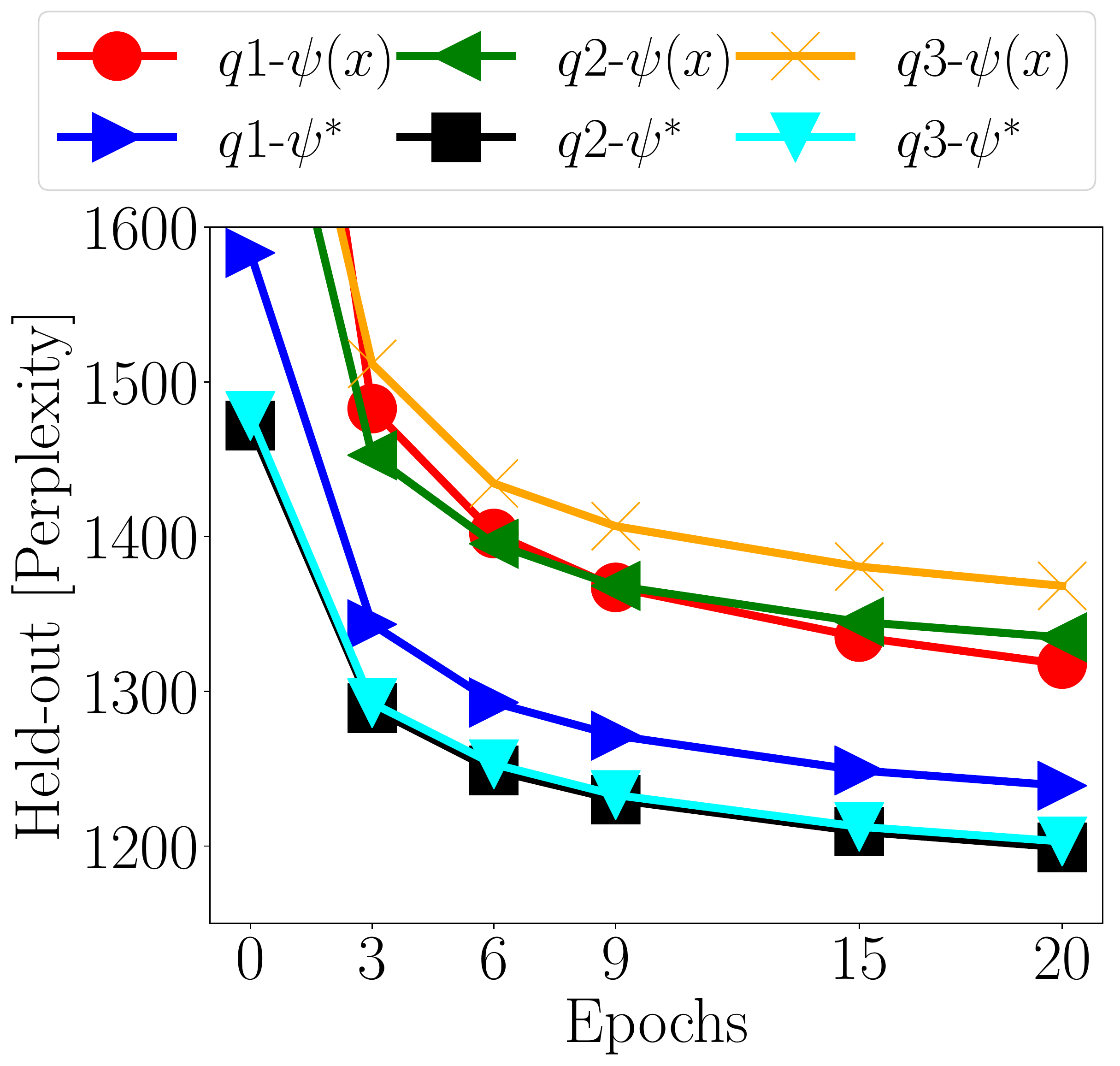}
	\caption{Held-out Data}
\label{fig:qvary_eval_bounds}
\end{subfigure}
\caption{
\small
\textbf{Varying the Depth of $\qzx$: }
Fig. \ref{fig:anneal_train_bounds} (\ref{fig:anneal_eval_bounds}) denotes the train (held-out)
perplexity for a three-layer generative model learned with inference networks of varying depth. 
The notation \Model{q3}{$\psi^*$} denotes that the inference network contained a two-layer 
intermediate hidden layer $h(x)=\text{MLP}(x;\phi_0)$ 
followed by $\mu(x) = W_{\mu}h(x),\log\Sigma(x)=W_{\log\Sigma}h(x)$.  
}
\label{fig:qvary_curves}
\end{figure*}

Can the overall effect of the additional optimization be \emph{learned} by the inference network at training time?
The experimental evidence we observe in Fig. \ref{fig:qvary_curves} suggests this is difficult.

When learning with $\local$, increasing the number of layers in the inference network
slightly decreases the quality of the model learned. 
This is likely because the already stochastic gradients of the inference
network must propagate along a longer path in a deeper 
inference network, slowing down learning of the parameters $\phi$ which in turn
affects $\local$, thereby reducing the quality of the gradients used to updated $\theta$. 

\section{Inference on documents with rare words}

\begin{figure*}[t!]
\centering
\begin{subfigure}{.3\linewidth}
  \centering
  \includegraphics[width=\linewidth]{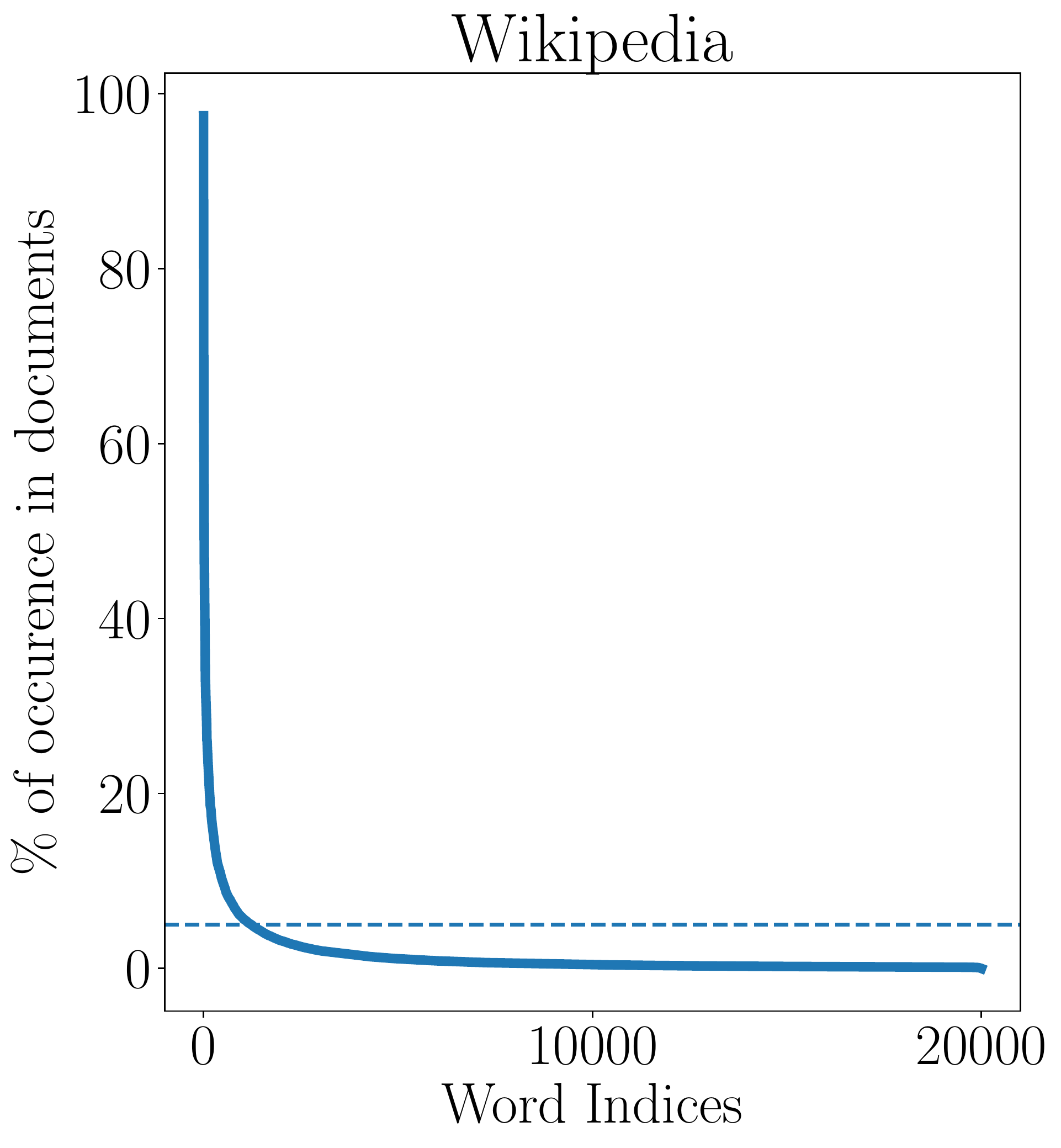}
  \caption{Sparsity of Wikipedia}
  \label{fig:wiki_sparse}
\end{subfigure}
\begin{subfigure}{.3\linewidth}
  \centering
  \includegraphics[width=\linewidth]{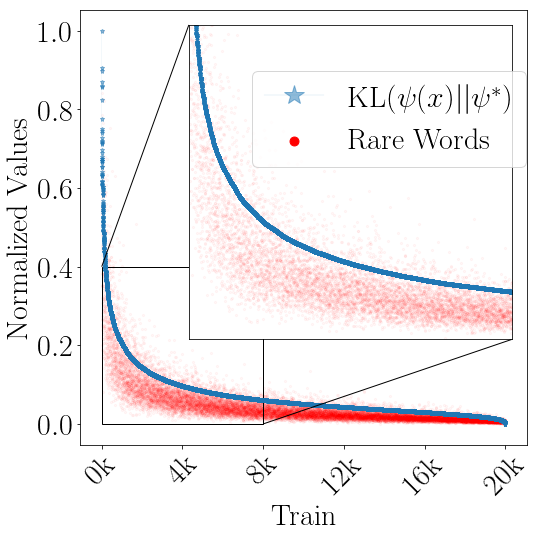}
  \caption{Training Data}
	\label{fig:norm_kl_train}
\end{subfigure}
\begin{subfigure}{.3\linewidth}
	\includegraphics[width=\linewidth]{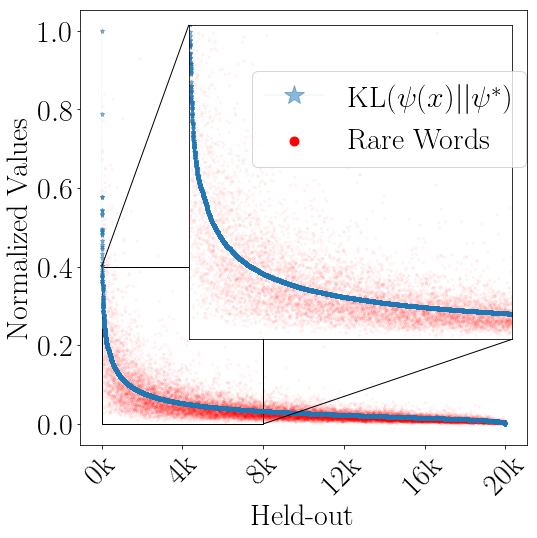}
	\caption{Held-out Data}
	\label{fig:norm_kl_eval}
\end{subfigure}
\caption{
\small
\textbf{Normalized KL and Rare Word Counts: }
Fig. \ref{fig:wiki_sparse} depicts percentage of times words appear in the Wikipedia dataset (sorted by frequency). The dotted
line in blue denotes the marker for a word that has a 5\% occurrence in documents. 
In Fig. \ref{fig:norm_kl_train}, \ref{fig:norm_kl_eval}, we superimpose (1) the normalized 
(to be between $0$ and $1$) values of $\text{KL}(\local \| \localopt)$ 
and (2) the normalized number of rare words (sorted by value of the KL-divergence)
for $20,000$ points (on the x-axis) randomly sampled from the train and held-out data. 
}
\label{fig:normalized_kl_vs_rare}
\end{figure*}

Here, we present another way to visualize the results of Fig. \ref{fig:kl_vs_rare}. 
We sample $20000$ training and held-out data points; we compute $\text{KL}(\local \| \localopt)$ (both are Normal distributions and the KL
is analytic) and the number of rare words in each document (where a word is classified as being rare if it occurs in less than $5\%$ of training documents).
We scale each value to be between $0$ and $1$ using: $\frac{c_i - min(c)}{max(c)-min(c)}$ where $c$ is 
the vector of KL divergences or number of rare words. We sort the scaled values by the KL divergence and plot them in Fig.
\ref{fig:normalized_kl_vs_rare}. As before, we observe that the documents that we move the farthest in KL divergence 
are those which have many rare words.

\end{document}